\documentclass[11pt,letterpaper]{article}
\usepackage[margin=1in]{geometry} 
\usepackage{graphicx}
\usepackage{amsmath}
\usepackage{multirow}
\usepackage{booktabs} 
\usepackage{array}  
\usepackage{makecell}
\usepackage{float}
\usepackage{hyperref}
\usepackage{verbatim}
\usepackage{microtype}
\usepackage{subcaption}
\usepackage{booktabs}
\usepackage{diagbox}
\usepackage{cuted}
\usepackage{longtable}
\usepackage{capt-of}
\usepackage{wrapfig}

\usepackage[utf8]{inputenc} % allow utf-8 input
\usepackage[T1]{fontenc}    % use 8-bit T1 fonts
\usepackage{hyperref}       % hyperlinks
\usepackage{url}            % simple URL typesetting
\usepackage{booktabs}       % professional-quality tables
\usepackage{amsfonts}       % blackboard math symbols
\usepackage{nicefrac}       % compact symbols for 1/2, etc.
\usepackage{microtype}      % microtypography
\usepackage{xcolor}         % colors

\title{Self-Creative Text-to-Object Generation using \\ Semantic-Aware Spatial Weighting}

\date{} 

\author{%
  Yue Yu\textsuperscript{1},
  Haibo Chen\textsuperscript{1},
  Shuo Chen\textsuperscript{2},
  Jian Yang\textsuperscript{1},
  Jun Li\textsuperscript{1}\thanks{Corresponding author.} \\[0.3em]
  \textsuperscript{1}Nanjing University of Science and Technology \quad
  \textsuperscript{2}Nanjing University \\[0.3em]
  \texttt{\{yuue, hbchen, csjyang, junli\}@njust.edu.cn}, \texttt{shuo.chen@nju.edu.cn}
}
\begin{document}

\maketitle

\begin{figure*}[ht]
\centering
\includegraphics[width=1\textwidth]{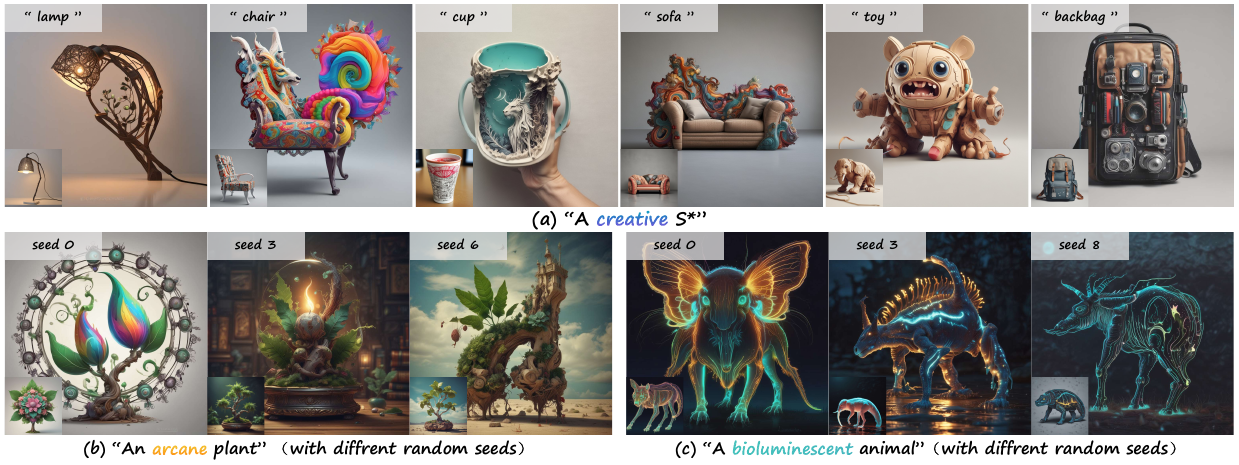}
\vskip -0.03in
\captionof{figure}{\textbf{Generating Creative Objects by our SCDiff}. (a) shows practical yet innovative designs using the prompt: \textit{“A creative S*"}. (b) and (c) demonstrate its ability to generate distinct conceptual objects within a broad category by varying the random seed.}
\label{fig:first}
\vskip -0.05in
\end{figure*}
\begin{abstract}
Instilling creativity in text-to-image (T2I) generation presents a significant challenge, as it requires synthesized images to exhibit not only visual novelty and surprise, but also artistic value. Current T2I models, however, are largely optimized for literal text-image alignment with their data distribution, and their noise prediction networks constrain the generation to high-probability regions, consequently generating outputs that lack authentic creativity. To address this, we propose a \textbf{S}elf-\textbf{C}reative \textbf{D}iffusion (\textbf{SCDiff}) model for meaningful T2I generations featuring two core modules: a learnable spatial weighting (LSW) module and a visual-semantic mixing loss (VSML). The LSW module designs a parametric Kaiser-Bessel window to reinforce central image features, fostering novel and surprising generation. The VSML module introduces a dual loss function: a similarity loss constrains that the new images align with its textual description, while a diversity loss maximizes its distinction from the original image, enhancing both semantic value and visual novelty. Extensive experiments demonstrate that our model substantially improves creativity, semantic alignment, and visual coherence, offering a simple yet powerful framework for generating creative objects. 
\end{abstract}

\vspace{-3mm}
\section{Introduction}

\label{sec:intro}
The rapid advancement of text-to-image (T2I) generation technologies \cite{VAE,DALL-E,StableDiffusion} has revealed their vast potential across diverse fields including digital art, game design, film production and advertising \cite{3D,Art_Education,survey,Game_Designers}.
Consequently, the generation of creative content has gained significant attention \cite{Conceptlab,TP20,Xiong2024NovelOS,CreativeAdversarialNetwork,C3}, as applications increasingly demand the production of new and meaningful visual objects. A central challenge in this pursuit is generating creative (\textit{e.g., novel, surprising, and valuable} \cite{boden2004creative}) objects.

While most T2I diffusion models \cite{StableDiffusion,SDXL-Lightning,sdxl_turbo,DDPM,SDXL} can generate images that align with simple category texts, they often fail to produce creative objects \cite{taxonomy,Role_Bias,understanding}, even when prompts include words like ``creative'' or more complex descriptors. This stems from the standard denoising process, which inherently converges toward high-density regions (\textit{i.e.}, common samples) of the training data distribution, resulting in conventional and homogenized outputs.

\setlength{\columnsep}{8pt}
\begin{wrapfigure}[17]{r}{0.5\textwidth}
% \begin{wrapfigure}{r}{0.5\textwidth}
  \begin{center}
  \vspace{-15pt}
    \includegraphics[width=0.48\textwidth]{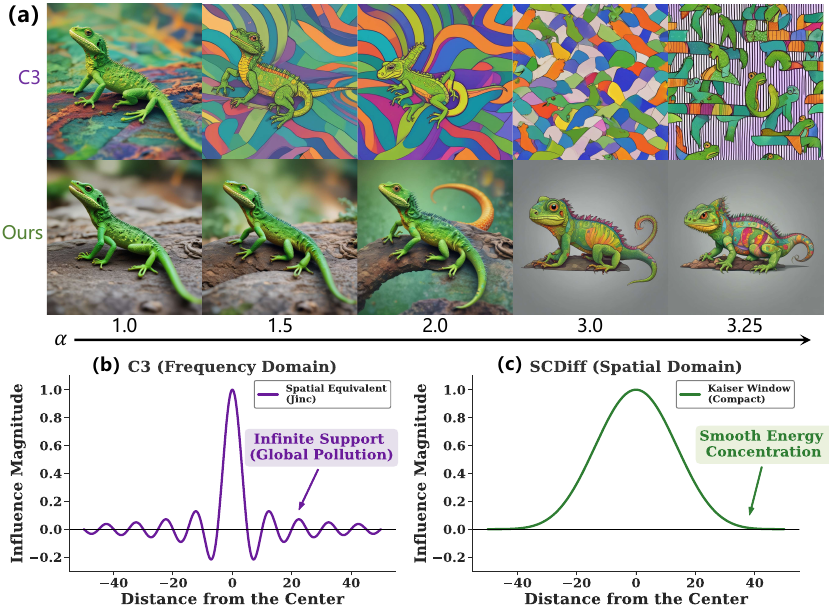}
  \vspace{-10pt}
  \end{center}
  \caption{Comparison of \textbf{SCDiff} (Ours) against existing methods (e.g., \textbf{C3} \cite{C3}). Increasing the amplification factor in C3 introduces ghosting artifacts.}
  \label{fig:c3-ghosting}
\end{wrapfigure}

One notable approach aimed at mitigating this issue is the Creative Concept Catalyst (C3) method \cite{C3}. C3 operates by transforming intermediate U-Net features \cite{U-Net} into the Fourier domain using the Fast Fourier Transform (FFT) \cite{bracewell2000fourier}, selectively amplifying low-frequency components with a fixed window to foster creative elements. 
However, C3 suffers from an inherent mathematical limitation: according to the convolution theorem \cite{gonzalez2018digital}, applying a hard cutoff in the frequency domain is equivalent to convolving the feature map with a Jinc function. As illustrated in Figure~\ref{fig:c3-ghosting} with the prompt \textit{“a colorful green lizard”}, this function possesses infinite spatial support characterized by oscillating side lobes. Consequently, the editing operation cannot be spatially localized, manifesting as severe global noise pollution in background regions.

To overcome these issues, we propose a \textbf{Self-Creative Diffusion (SCDiff)} model to effectively produce creative objects. Guided by the three classical characteristics of creativity—novelty, surprise, and value \cite{boden2004creative}, our model adaptively enhances spatial features to boost novelty and surprise, and simultaneously leverages textual semantics to maintain meaningfulness. Specifically, SCDiff comprises two modules. The \textbf{Learnable Spatial Weighting (LSW)} module employs a learnable Kaiser-Bessel window~\cite{kaiser1974non,dobson1994advanced} to adaptively reinforce central image features. While the radius $R$ is user-defined to constrain the editing extent, the module dynamically adjusts the window profile parameter $\beta$ and amplification factor $\alpha$. This induces a significant distribution shift in the central region, effectively disrupting the model's tendency to converge towards the mean patterns of the training distribution. Constrained by the diversity and similarity losses, the model is guided to explore low-probability yet semantically plausible sub-manifolds, thereby generating outputs with enhanced novelty and surprise. A \textbf{Visual-Semantic Mixing Loss (VSML)} module introduces a dual-loss mechanism to balance the output. A similarity loss constrains the generated image to align with its textual description, ensuring semantic value and preventing meaningless outputs from the LSW. A diversity loss increases visual distinction from the original image, thereby explicitly promoting novelty. 

Our contributions are summarized as follows: \textbf{(1)} We propose a learnable spatial weighting method that performs controllable feature enhancement directly in the spatial domain. Its tunable parameters enable the model to accentuate target regions while maintaining global structural consistency. \textbf{(2)} We introduce a semantic constraint module that adjusts the amplification strength based on feature-level novelty scores, enabling a balance between visual novelty, surprise, and semantic meaningfulness. \textbf{(3)} Through extensive experiments on multiple T2I baselines, we demonstrate that our method significantly improves semantic alignment, creativity, and visual coherence, and concurrently mitigates ringing artifacts, resulting in a powerful and robust solution for enhancing creativity.

\begin{figure*}[t]
\centering
%\vspace{-10mm} 
\includegraphics[width=0.97\textwidth]{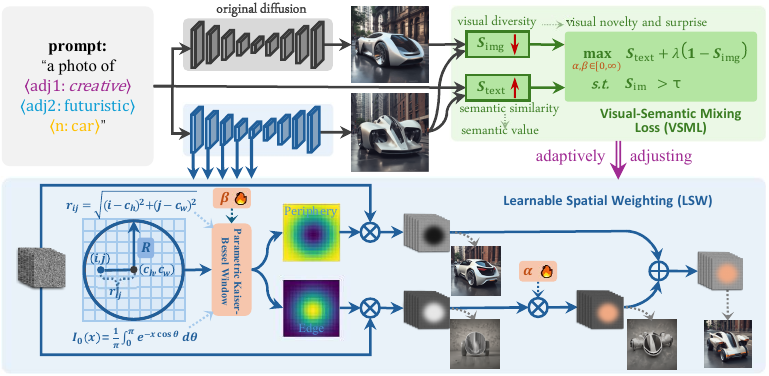}
\vspace{-2mm}
\caption{Overview of our SCDiff. It advances diffusion models through two novel modules: (1)  a parametric Kaiser-Bessel window (LSW) modulating U-Net encoder features, and (2) a dual VSML loss that optimizes these parameters for self-creative generation.}
\label{fig:framework}
\vspace{-3mm}
\end{figure*}

\vspace{-3mm}
\section{Related Work}

Recent diffusion models, such as Imagen \cite{imagen}, DALLE-2 \cite{DALL-E}, and Stable Diffusion \cite{sdxl_turbo, SDXL-Lightning, StableDiffusion}, have achieved unprecedented levels of image synthesis quality, shifting the focus toward enhancing the diversity and creativity of generated images. Meanwhile, Large Language Models \cite{openai2024gpt4o} enhance text-to-image (T2I) generation by leveraging its advanced language understanding, providing creative guidance to produce images that better align with the user’s intent.

Recent studies focus on combinational visual creativity, generating novel objects by merging features from multiple concepts into a unified and coherent representation \cite{ELS,Inspiration_Tree,ConceptCraft,TRTST,seu-RedefiningID}. PartCraft \cite{PartCraft} allows for the flexible composition of parts from diverse objects, producing novel yet holistically plausible results that maintain structural coherence. TP2O \cite{TP20,AGSwapOC} introduces a balanced swap-sampling strategy to creatively combine textual descriptions of two objects. ATIH \cite{Xiong2024NovelOS} employs a balanced loss function along with a similarity score to fuse object texts with images, generating new distinctive objects.

Concurrently, another category of innovation methods are dedicated to enhancing creativity within the boundaries of given categories \cite{Conceptlab,Yu2025DMFFT,ProCreate}. ConceptLab \cite{Conceptlab} combines vision-language models with diffusion priors to drive the creation and hybrid exploration of novel concepts.  C3 \cite{C3} enhances novelty and unconventional features by selectively amplifying intermediate activations during the denoising process, without requiring additional training. In contrast, we propose a feature enhancement paradigm based entirely on the spatial domain. We utilize a parameterized Kaiser-Bessel window to perform precise local modulation directly on intermediate features. This method effectively stimulates creative features while avoiding spectral aliasing and ringing artifacts caused by frequency truncation.

\vspace{-3mm}
\section{Self-Creativity Diffusion Model}
\label{sec:method}
In this section, we introduce a \textbf{Self-Creativity Diffusion (SCDiff)} model for generating creative objects, as illustrated in Fig. \ref{fig:framework}. Our model is built upon two key components: a \textit{Learnable Spatial Weighting} (\textbf{LSW}) module that applies an adaptive Kaiser–Bessel window to shallow U-Net features, and a \textit{Visual-Semantic Mixing Loss} (\textbf{VSML}) module that combines a similarity loss enforcing text alignment with a diversity loss encouraging deviation from the original image, jointly enhancing semantic consistency and visual novelty.

\vspace{-3mm} 
\subsection{Learnable Spatial Weighting}
Creativity research formalizes generation as search within a constrained conceptual space~\cite{boden2004creative}. In the Creative Systems Framework (CSF)~\cite{wiggins2006searching}, the conceptual space $\mathcal{C}$ is governed by a rule set $\mathcal{R}$ and defined as the subset of a universal space $\mathcal{U}$ satisfying an acceptability condition:\begin{equation}
\mathcal{C} = \{c \mid c \in \mathcal{U} \wedge \mathcal{A}(\mathcal{R}, c) \text{ is true}\},
\end{equation}
where $\mathcal{A}$ denotes the acceptability operator. The transition from an existing concept $c_{\text{in}}$ to a novel concept $c_{\text{out}}$ is realized by a Traversal operator $\mathcal{T}$ and an Evaluation operator $\mathcal{E}$:\begin{equation}
c_{\text{out}} = \mathcal{T}(\mathcal{R}, \mathcal{E})(c_{\text{in}}).
\end{equation}
We instantiate the CSF framework within the latent manifold of a diffusion model. Let the intermediate feature maps $\mathbf{x}$ of the denoising U-Net (or DiT blocks) constitute the universal space $\mathcal{U}$. To induce structural novelty, we propose an LSW module that modulates the encoder feature $\mathbf{x}$ spatially to operationalize the rule set $\mathcal{R}$, while $\alpha$ acts as the traversal operator $\mathcal{T}$ to drive generation from the original concept $c_{\text{in}}$ ($I_{\text{org}}$) to novel structural variants $c_{\text{out}}$.

Given a structured prompt template $T_p$—\textit{``a photo of  $<adj_1:$ creative or $adj_2>$ $<$object name$>$.''}—we extract feature maps $\mathbf{x} \in \mathbb{R}^{B \times C \times H \times W}$ from a single encoder block (selected from the set $\{\textit{Down}_0, \textit{Down}_1, \textit{Down}_2, \textit{Mid}\}$). Note that upsampling layers are excluded to avoid high-frequency noise (see Appendix \ref{sec:feature_enhancement_discussion}). Using the $T_p$, the base diffusion model first generates an original image $I_{\text{org}}$. These modulated features are then passed via skip connections to the decoder, generating diverse objects.

To realize the traversal operator $\mathcal{T}$ within the CSF framework, we introduce spatial feature modulation; however, traditional hard-threshold windows with fixed radius \cite{C3} often introduce abrupt truncation, causing boundary discontinuities and spectral ringing, as illustrated in Fig. \ref{fig:c3-ghosting}. To mitigate this, our LSW module implements a parametric smooth Kaiser–Bessel window for large‑scale feature modulation. Formally, the operation is defined as:
\begin{align}
\mathbf{x}^{\text{KB}}(\alpha,\beta)= \mathbf{x} \odot (1 - \mathbf{w}(\beta)) + \alpha \cdot \mathbf{x} \odot \mathbf{w}(\beta),
\label{eq:KBoperation}
\end{align}
where $\mathbf{w}(\beta) =\textsc{Repeat}(w(\beta), BC)$ replicates the window $w(\beta)$ across $BC$ feature maps, $\alpha$ is a learnable amplification factor, and $\beta$ controls the window shape. The window $w(\beta)\in \mathbb{R}^{H \times W}$ is: 
\begin{align}
w_{ij}(\beta) =
\begin{cases}
\displaystyle 
\frac{I_0\!\left( \beta \sqrt{1 - (r_{ij}/R)^2} \right)}{I_0(\beta)}, & r_{ij} \le R, \\[6pt]
0, & r_{ij} > R,
\end{cases}
\label{eq:wl}
\end{align}
where  $r_{ij} = \sqrt{(i - H/2)^2 + (j - W/2)^2}$ denotes the radial distance of pixel $(i,j)$ from the image center, $R$ is a cutoff radius hyper-parameter, and $I_0(\cdot)$ is the zeroth‑order modified Bessel function of the first kind, $
I_0(x) = \frac{1}{\pi} \int_0^\pi e^{-x \cos(\theta)} \, d\theta$ \cite{watson1995treatise}. The smooth tapering of $w_{\beta}$ provides a differentiable spatial weighting for feature maps, allowing adjustment of large‑scale structures.

% \begin{figure*}[t]
% \centering

% %\framebox[4.0in]{$\;$}
% \includegraphics[width=0.87\textwidth]{images/17_window_comparision.pdf}
% \vskip -0.05in
% \caption{Results using different spatial window profiles. Note that the Gaussian window (b), despite its smooth appearance, fails to trigger creative geometric deformation due to its overly-concentrated (peaked) energy distribution. In contrast, our Kaiser-Bessel window (d) facilitates holistic structural reorganization by providing a sustained flat-top energy profile while ensuring strict boundary compactness.}
% \label{fig:window_comparison}
% \vskip -0.1in
% \end{figure*}

% \subsection{Theoretical Analysis}
% \textbf{Discussion 1: Spatial-Domain vs. Frequency-Domain.} 
The rationale for this spatial-domain design can be justified by the theoretical limitations of frequency-domain structure isolation. \textit{Frequency‑domain editing methods} (e.g., C3~\cite{C3}) typically isolate the structure by applying a circular low‑pass filter (mask $M$) in the frequency domain. However, by the convolution theorem \cite{gonzalez2018digital}, a sharp circular cutoff in frequency corresponds in the spatial domain to convolution with a \textit{Jinc function}:
\begin{equation}
\mathcal{F}^{-1}{M(u,v)} \propto J_1(2\pi W r)/r,
\end{equation}
where $J_1$ is the first-order Bessel function \cite{watson1995treatise}, $W$ denotes the spectral cutoff frequency, and $r$ represents the spatial radial distance. This Jinc function has \textbf{infinite spatial support} and exhibits oscillating side‑lobes (“ringing”) that decay asymptotically but never vanish completely (Fig.~\ref{fig:c3-ghosting}(b)). Crucially, the spatial support of this kernel is infinite:
\begin{equation}
\text{Supp}(\mathcal{F}^{-1}{M}) = { r \mid r \in [0, \infty) },
\end{equation}
As a result, frequency‑domain operations inherently affect the entire spatial domain, causing semantic drift in background regions (C3 in Fig.~\ref{fig:c3-ghosting}(a)). In contrast, \textit{Spatial-Domain methods} like our Kaiser‑Bessel window $w(\beta)$ are \textbf{compactly supported}, satisfying the following boundary condition: $w_{ij}(\beta) \equiv 0, \quad \forall r_{ij} > R$. This mathematical locality guarantees that $\mathbf{x}^{\text{KB}}(\alpha,\beta) \equiv \mathbf{x}$ outside the local region of interest (Fig.~\ref{fig:c3-ghosting}(c)). This strict locality mathematically ensures that background pixels remain unperturbed, thereby eliminating the global noise and “ghosting” artifacts characteristic of frequency‑based methods (ours in Fig.~\ref{fig:c3-ghosting}(a)).

\setlength{\columnsep}{8pt}
\begin{wrapfigure}[12]{r}{0.45\textwidth}
% \begin{wrapfigure}{r}{0.5\textwidth}
  \begin{center}
  \vspace{-25pt}
    \includegraphics[width=0.45\textwidth]{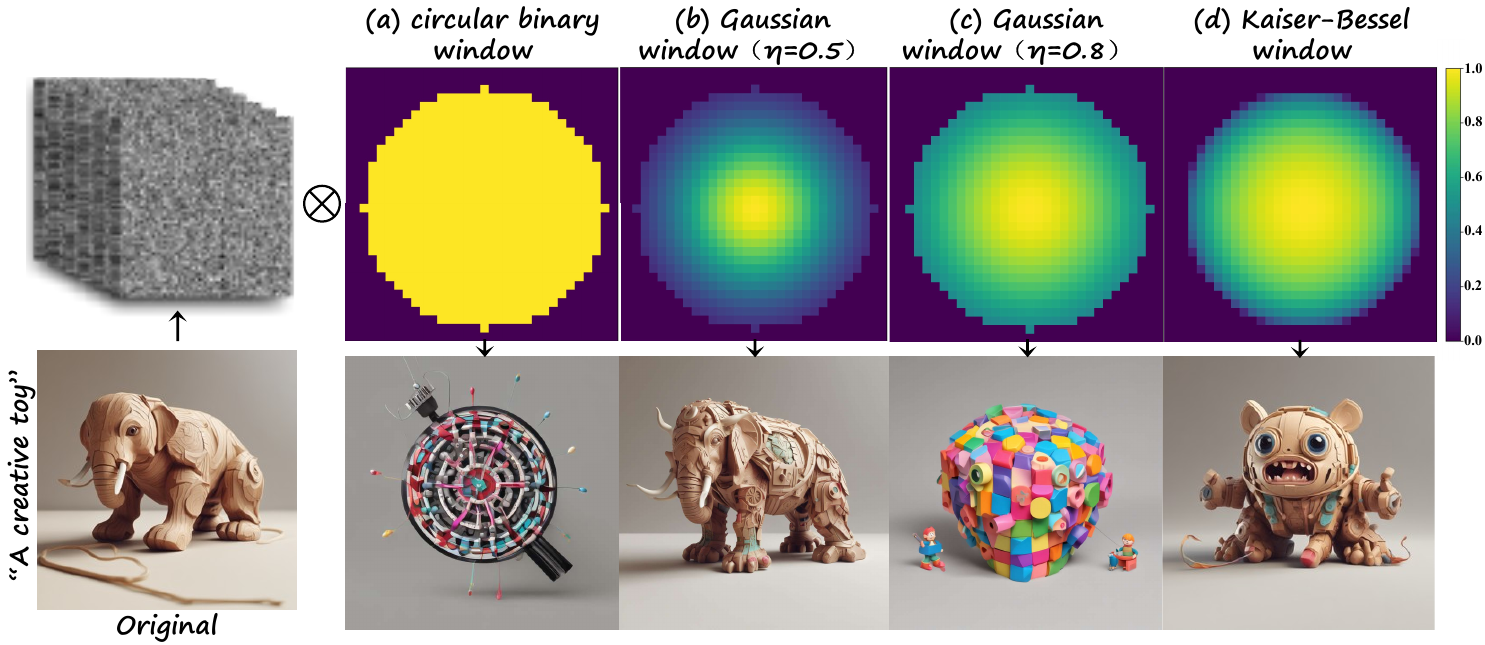}
  \vspace{-30pt}
  \end{center}
  \caption{Results using different spatial window profile: (a) circular binary, (b) Gaussian ($\eta$=0.5), (c) Gaussian ($\eta$=0.8), and (d) our Kaiser-Bessel window.}
  \label{fig:window_comparison}
\end{wrapfigure}
% \textbf{Discussion 2: Kaiser-Bessel Window Selection.} 
While compact support eliminates global artifacts, the profile within $r_{ij}\le R$ determines whether the decoder interprets the modulation as geometric transformation or mere texture variation. We therefore examine three candidate windows: (1) A naive \textit{circular binary window} creates a sharp discontinuity that the diffusion model often misinterprets as a semantic boundary, manifesting as unnatural radial discontinuities that prevent the object from blending seamlessly with the background (Fig. \ref{fig:window_comparison}(a)). (2) \textit{Gaussian window} provide a smooth baseline but suffer from an inherent deficiency in energy-localization efficiency: to ensure a gradual decay to zero at the boundary $R$, a small width $\eta$ is required, which concentrates amplification in a narrow central zone(Fig. \ref{fig:window_comparison}(b)). The decoder typically interprets this localized energy peak as texture variation rather than geometric transformation. Conversely, increasing $\eta$ to flatten the profile reintroduces boundary discontinuities and thus boundary artifacts (Fig. \ref{fig:window_comparison}(c)). (3) \textit{Kaiser‑Bessel window}, which approximates Discrete Prolate Spheroidal Sequences (DPSS), in contrast, is designed to maximize energy concentration within a finite interval \cite{kaiser1974non,harris1978use}. By adjusting the shape parameter $\beta$, the central profile can be broadened while maintaining a rigorously smooth transition to zero. This shape forces the decoder to focus on global structural reorganization rather than local texture changes (Fig. \ref{fig:window_comparison}(d)).

\begin{figure*}[t]
\centering
%\framebox[4.0in]{$\;$}
\includegraphics[width=0.97\textwidth]{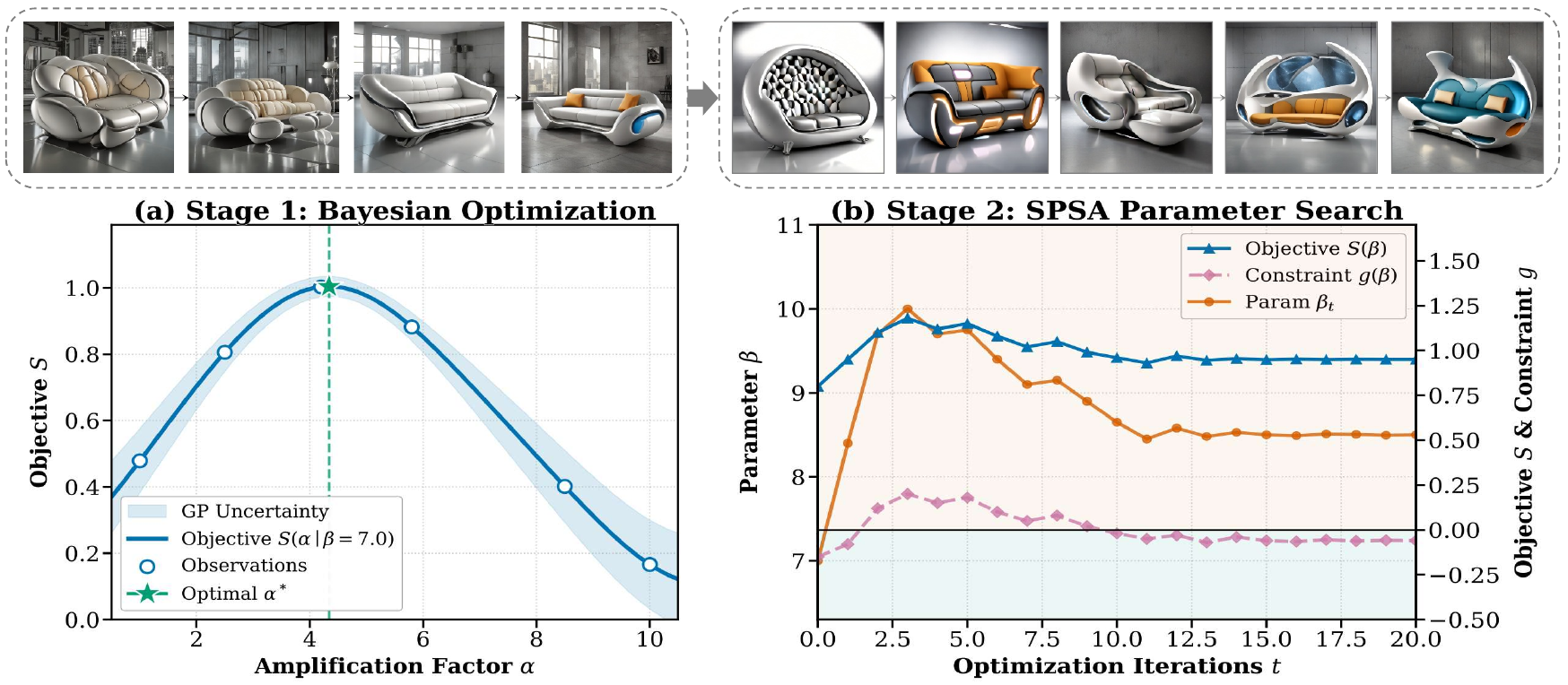}
\vskip -0.05in
\caption{Visualizing the updated process of our VSML using the prompt \textit{'a futuristic sofa'.}}
\vspace{-15pt}
\label{fig:iteration}
\end{figure*}

% \textbf{Discussion 3: Energy-Geometry Coupling and the Creative Manifold}

%, denoted as the reconstructed image $I_{\text{rec}}$ J(\boldsymbol{\theta}) = \underbrace{S(T_p, I(\alpha,\beta))}_{\text{maximize semantic similarity for value }} + \lambda \cdot \underbrace{\left( 1 - S(I_{\text{org}}, I(\alpha,\beta)) \right)}_{m},
\vspace{-3mm}
\subsection{Visual-Semantic Mixing Loss}
The LSW module employs a tunable Kaiser–Bessel window (Eq.~\ref{eq:KBoperation}) for spatial modulation, parameterized by the amplification factor $\alpha$, the shape parameter $\beta$, and the user-defined hyperparameter $R$. Its centroid $(x_c, y_c)$ defaults to the image center but is adjustable for localized off-center enhancement. This module generates diverse object images, denoted as $I(\alpha,\beta)$, through a complete denoising trajectory. To dynamically optimize $\alpha$ and $\beta$—thereby allowing the model to automatically balance energy distribution and spectral smoothness—we introduce a Visual-Semantic Mixing Loss (VSML) in this subsection, which serves as the evaluation operator $\mathcal{E}$ in the CSF framework. The objective of VSML is twofold: first, to enhance the visual novelty and surprise by increasing the diversity between $I(\alpha,\beta)$ and the original image $I_{\text{org}}$; and second, to preserve semantic value by constraining $I(\alpha,\beta)$ using its similarity to the original text prompt $T_p$. Formally, VSML is defined as follows:\begin{equation}
\max_{\alpha,\beta} \ \mathcal{S}(\alpha,\beta), \quad \text{s.t.} \ \ g(\alpha,\beta) \le 0,
\label{eq:image_text_loss}
\end{equation}
where $\mathcal{S}(\alpha,\beta)= S_{\text{text}}(T_p, I(\alpha,\beta))+ \lambda \cdot \left( 1 - S_{\text{img}}(I_{\text{org}}, I(\alpha,\beta) ) \right)$, $g({\beta}) = \tau - S_{\text{img}}(I_{\text{org}}, I(\alpha,\beta))$, $S(\cdot, \cdot)$ represents the CLIP \cite{CLIP} cosine similarity, where lower similarity to the original generated image indicates higher novelty following Elgammal and Saleh~\cite{Elgammal2015QuantifyingCI}, $\tau$ acts as a safety margin and is set to $\tau = 0.7$. This constraint ensures the model does not mistakenly reward severely distorted, low-similarity images with high creative scores. While the VSML appears simple, empirically finding parameters for creative generation is challenging when simultaneously adjusting  $\alpha$ and $\beta$ as $\alpha$ varies widely. Here, we introduce a hierarchical adjustment strategy to learn the parameters $\alpha, \beta$. The procedure proceeds as follows. 
First, with the shape parameter $\beta$ fixed at its empirically determined value ($\beta = 7.0$), we consider the VSML loss $\mathcal{S}(\alpha,\beta)$ without imposing the constraint $g(\alpha,\beta) \le 0$. Given the unimodal yet expensive-to-evaluate response of $\alpha$ observed in Fig. \ref{fig:iteration}, we employ Bayesian optimization~\cite{Snoek2012PracticalBO} to efficiently determine the optimal amplification factor $\alpha^{*}$:\begin{equation}
\alpha^{*} = \arg\max_{\alpha \in [1.5,, 10.0]} \mathcal{S}(\alpha \mid \beta=7.0).
\end{equation}
Since each query of $\alpha$ requires a complete diffusion forward pass, we employ a Gaussian Process surrogate model with the Expected Improvement (EI) acquisition function to conduct sample-efficient optimization.

Next, using the resulting $\alpha^{*}$ together with the user-specified $R$, we refine the window profile parameter $\beta$ by solving the constrained optimization problem:\begin{equation}
\max_{\beta} \ \mathcal {S}(\beta \mid \alpha^*), \  \  s.t. \ \ g(\beta \mid \alpha^*) \le 0.
\label{eq:spsa-problem}
\end{equation}
This problem is solved using the Simultaneous Perturbation Stochastic Approximation (SPSA)~\cite{spall1998implementation}, a zeroth-order method that estimates pseudo-gradients via simultaneous random perturbations. Further details for Bayesian optimization and SPSA are consolidated in Appendix \ref{sec:Hierarchical}.
Fig.~\ref{fig:iteration} illustrates this hierarchical optimization process. By strictly adhering to the user-defined spatial constraint ($R=15$), this adaptive mechanism automatically identifies the optimal combination of intensity and window profile, ensuring the generation of semantically consistent and structurally reliable creative images.

\vspace{-3mm}
\section{Experiments}

\vspace{-3mm}
\subsection{Experiment Settings}
\label{sec:setting}
\textbf{Datasets.} We comprehensively extend the COF dataset~\cite{AGSwapOC} in two ways. First, we build a simple variant-\textit{CreaCOF}-by directly appending the prefix ``creative'' to each of the 950 original class names. Second, we construct another variant-\textit{AdjCOF}-by assigning 10 unique and tailored adjectives to each of the 95 superclasses, as detailed in  \textbf{Appendix} \ref{sec:datasets}. This yields 9,500 prompt variants with more complex syntactic structures and increased generative difficulty.

\textbf{Implementation Details.} We selected SDXL-Lightning~\cite{SDXL-Lightning} (1-step version) and SDXL-Turbo~\cite{sdxl_turbo} for quantitative and qualitative evaluation. All experiments were conducted on an NVIDIA RTX 3090 GPU to ensure consistent and reproducible results. Our framework performs an automated parameter search (approximately 38 seconds) to determine the optimal configuration. Once the parameters are identified, the resulting inference latency is minimal (2–3 seconds).

%We selected two different Stable Diffusion base models for both quantitative and qualitative analysis: Lightning~\cite{SDXL-Lightning} (1-step version), and Turbo~\cite{sdxl_turbo}. All experiments were conducted on an NVIDIA RTX 3090 GPU to ensure the efficiency and accuracy of the results. Our framework performs an automated parameter search (approx. 4 mins) to find the optimal configuration. Once parameters is identified, the inference latency is negligible (2-3 seconds)
% A full run of the pipeline takes approximately 4 to 5 minutes, depending on the number of adaptive rounds.
% Our source code will be made publicly available.

\textbf{Evaluation Metrics.} We evaluate our method along two primary dimensions: \textit{\textbf{Semantic Alignment (SA)}} and \textit{\textbf{Creativity}}. 
For \textit{\textbf{SA}}, we assess text-image alignment with CLIP similarity~\cite{CLIP} and object-level accuracy via LLaVA-VQA~\cite{llava}. To decouple creative divergence from image degradation, we additionally report P-tile~\cite{ptitle} as a quality indicator, confirming that the observed FID shift reflects structural novelty rather than synthesis artifacts. Following the C3 protocol~\cite{C3}, we measure \textit{\textbf{Creativity}} using Fréchet Inception Distance (FID) and Precision — higher FID and lower Precision indicate greater creativity. Diversity is assessed through three higher-is-better metrics: kNN-Coverage Recall in CLIP space~\cite{recall}, mean LPIPS distance in AlexNet space~\cite{lpips} , and the Vendi Score~\cite{Vendi}. We also evaluate creative merit via GPT-4o, scoring novelty, surprise, and value~\cite{boden2004creative}.

\begin{figure*}[h]
\centering
\vspace{-20pt}
\includegraphics[width=0.97\textwidth]{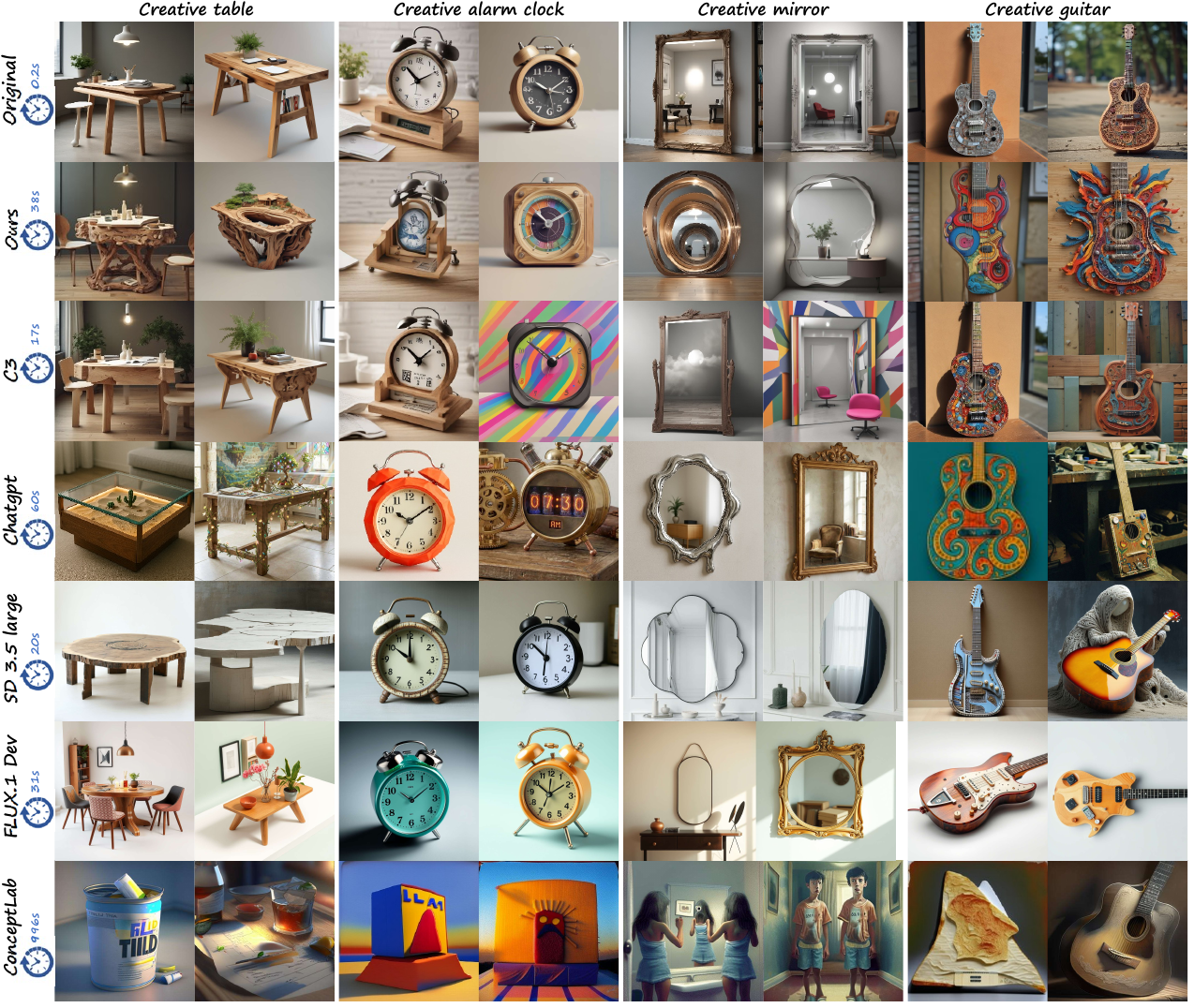}
\caption{Qualitative Comparisons on \textit{CreaCOF} demonstrate that our method delivers creative and meaningful outputs.}
\vspace{-20pt}
\label{fig:creative-comparable-result}
\end{figure*}

\begin{table*}[h]
\vspace{-8pt}
\centering
\setlength{\tabcolsep}{2pt}
\renewcommand{\arraystretch}{1.1}
\resizebox{0.97\linewidth}{!}{
\begin{tabular}{c|c|ccc|ccccc|cccc}
 \Xhline{1.2pt}
\multirow{2}{*}{Model} & \multirow{2}{*}{Method} & \multicolumn{3}{c|}{Semantic Alignment} & \multicolumn{5}{c|}{Creativity Metrics} & \multicolumn{4}{c}{Creativity Evaluated by GPT-4o} \\
\cline{3-14}
 & & CLIP$\uparrow$ & LLaVA$\uparrow$ & P-tile$\downarrow$ & FID*$\uparrow$ & Precision$\downarrow$ & Recall$\uparrow$ & LPIPS$\uparrow$ & Vendi$\uparrow$ & Surprise$\uparrow$ & Value$\uparrow$ & Novelty$\uparrow$ & Overall$\uparrow$\\
\hline
% Turbo-Easy
\multirow{3}{*}{\makecell{Turbo\\(CreaCOF)}} 
& Orig~\cite{sdxl_turbo} & 0.290& 0.742& 0.025
& 664.717 & 0.997 & 0.500 & 0.520 & 1.737 
& 5.4& 6.0& 5.8& 6.0\\
% \cline{2-14}
& C3~\cite{C3} & 0.295& 0.784& 0.034
& 703.850 & 0.958 & 0.690 & \textbf{0.577}& 2.016 
& 6.0& 6.2& 6.4& 6.4
\\
% \cline{2-14}
& Our SCDiff & \textbf{0.308}& \textbf{0.911}& \textbf{0.024}& \textbf{722.057}& \textbf{0.955}& \textbf{0.719}& 0.556 & \textbf{2.188}& \textbf{7.8}& \textbf{7.2}& \textbf{7.8}& \textbf{7.8}\\
\hline

% Lightning-Easy
\multirow{3}{*}{\makecell{Lightning\\(CreaCOF)}} 
& Orig~\cite{sdxl_turbo} & 0.282& 0.503& 0.027
& 879.685 & 0.942 & 0.178 & 0.525 & 2.096 
& 7.7& 8.2& 8.0& 8.0
\\
% \cline{2-14}
& C3~\cite{C3} & 0.287& 0.629& 0.021
& 1064.391 & 0.842 & 0.269 & \textbf{0.611}& 2.163 
& 7.5& 7.5& 7.8& 7.8
\\
% \cline{2-14}
& Our SCDiff & \textbf{0.290}& \textbf{0.743}& \textbf{0.012}& \textbf{1089.216}& \textbf{0.838}& \textbf{0.289}& 0.596 & \textbf{2.259}& \textbf{9.0}& \textbf{8.3}& \textbf{8.8}& \textbf{8.8}\\
\hline
% Reference models
\multirow{3}{*}{\makecell{Reference \\ (CreaCOF)}}
% \multirow{2}{*}{Comparison Methods \\(CreaCOF)} 

& SD3.5 Large~\cite{sd3.5} &  0.305 & 0.747 &  0.016
& 642.351  &  0.926 & 0.673  & 0.563 & 2.371 
& 8.2 & 7.5 & 8.0 & 8.0 \\
\cline{2-14}
& FLUX.1 Dev~\cite{flux} & 0.303 &  0.820 &  0.014
&  705.872 &0.945 & 0.579 & 0.541 & 2.085
& 7.7 & 7.5 & 7.8 & 7.7 \\
\cline{2-14}
% \hline
&ConceptLab~\cite{Conceptlab} & 0.220& 0.291& 0.018& 2025.712 & 0.341 & 0.375 & 0.679 & 2.673 & 8.1& 7.6& 8.2& 8.1\\

\hline\hline
% Turbo-AdjCOF
\multirow{3}{*}{\makecell{Turbo\\(AdjCOF)}} 
& Orig~\cite{SDXL-Lightning} & 0.292& 0.515& \textbf{0.022}& 649.324 & 0.980 & 0.442 & 0.547 & 1.716 
& 7.4& 7.9& 7.2& 7.7\\
% \cline{2-14}
& C3~\cite{C3} & 0.297& 0.612& 0.027
& 675.499 & 0.959 & 0.516 & 0.561 & 2.095 
& 6.8& 7.5& 6.8& 7.1
\\
% \cline{2-14}
& Our SCDiff & \textbf{0.305}& \textbf{0.658}& 0.024
& \textbf{789.587}& \textbf{0.952}& \textbf{0.565}& \textbf{0.579}& \textbf{2.113}& \textbf{8.7}& \textbf{8.3}& \textbf{8.8}& \textbf{8.7}\\
\Xhline{1.2pt}
\end{tabular}}

\caption{Comprehensive quantitative evaluation comparing novelty, diversity and usability metrics across different models and methods. To ensure that FID and Precision reflect genuine structural novelty rather than generation failures, we apply a uniform quality filter ($\text{P-tile} > 0.04$)~\cite{ptitle} to the outputs of all methods before evaluation.}
\label{tab:comprehensive_analysis}
\vskip -0.2in
\end{table*}

%  & SDXL-Lightning~\cite{SDXL} & 0.302& 0.559& 0.025
% & 583.105 & 0.961 & 0.500 & - & - 
% & 7.6& 7.7& 7.8& 7.8\\
% \cline{2-14}

\subsection{Main Results}
\vspace{-3mm}
To ensure a comprehensive evaluation of our SCDiff, we compare it against C3 \cite{C3}, ConceptLab \cite{Conceptlab}, along with relevant baseline models \cite{sdxl_turbo,SDXL-Lightning}. For a fair comparison, we also adaptively tuned the parameter $\alpha$ of C3.

\begin{figure*}[h]
\centering
%\framebox[4.0in]{$\;$}
\includegraphics[width=0.97\textwidth]{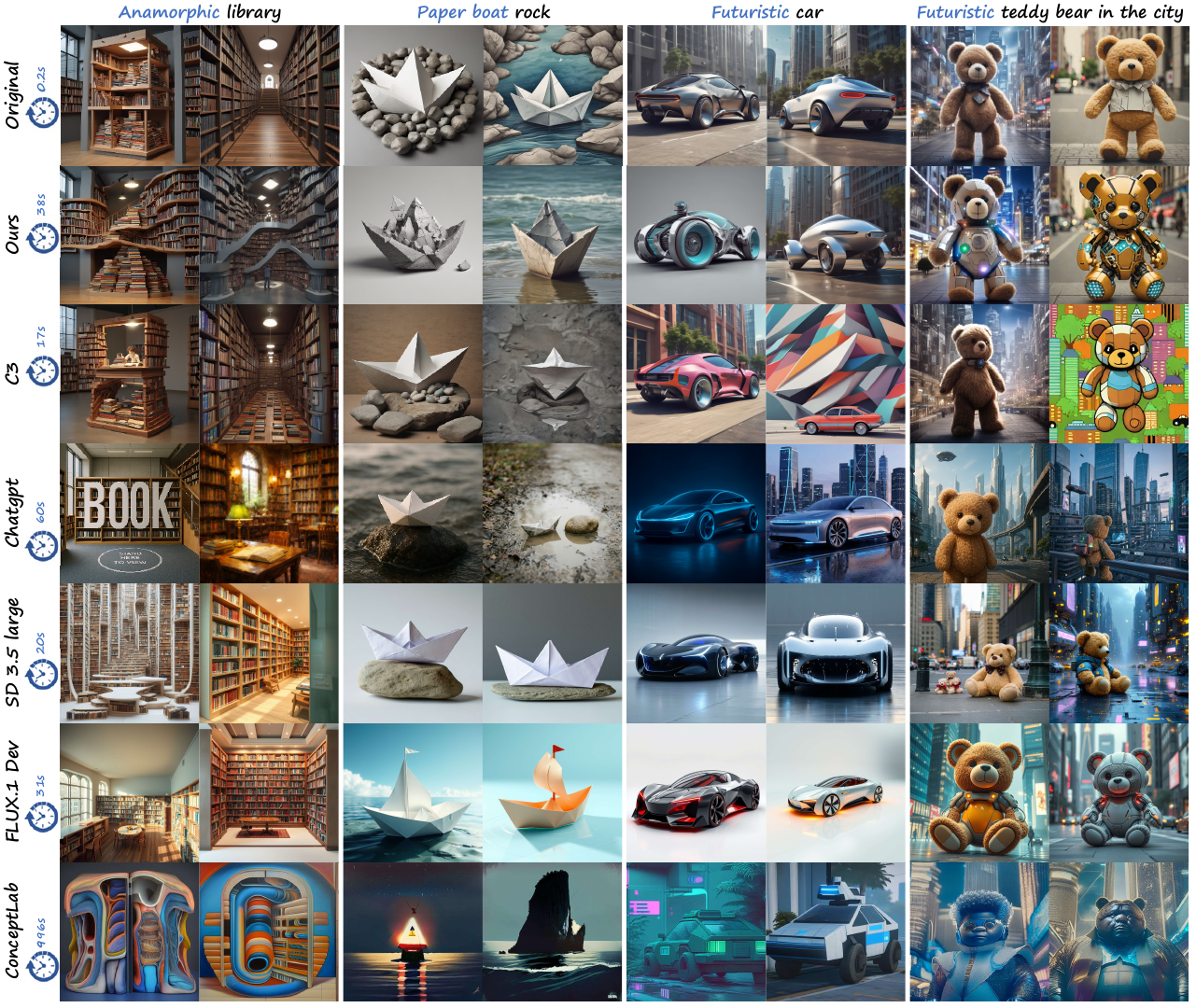}
\vskip -0.05in
\caption{Qualitative Comparisons on \textit{AdjCOF} demonstrate that our method consistently produces visually impressive results.}
\label{fig:adjectives-comparable-result}
\vspace{-15pt}
\end{figure*} %with Out-of-Distribution (OOD) Adjectives Handling.

\textbf{Quantitative Results.}
Table~\ref{tab:comprehensive_analysis} presents the quantitative results on both the CreaCOF and AdjCOF benchmarks, yielding three key observations: First, our SCDiff significantly outperforms all baselines on the CLIP-based semantic alignment score. This strong alignment directly translates to exceptional performance on the LLaVA-VQA metric, confirming that the generated images faithfully and clearly represent objects. Furthermore, our method attains a superior $\text{P-tile}$ score while minimizing distortion. Together, these results demonstrate our model’s unique ability to enhance semantic alignment without sacrificing visual fidelity.

Second, our SCDiff effectively balances creativity and fidelity. It attains a high FID score with only a controlled reduction in Precision. In contrast, ConceptLab’s excessively low Precision indicates substantial deviation and degraded fidelity. To ensure the rigor and reliability of our evaluation, we applied a filter ($\text{P-tile} > 0.04$) across all methods (including SCDiff and all competitive baselines) to mitigate the impact of low-quality artifacts, which could otherwise distort the FID and Precision metrics. Turbo-SCDiff achieves the highest Recall (0.719) among all compared methods, and its Vendi score (2.188) outperforms C3 and FLUX, though it trails SD3.5 Large (2.371). Its LPIPS (0.556) is slightly below C3 (0.577) but remains above the baseline (0.520). For Lightning-SCDiff, Recall (0.289) is lower than FLUX and SD3.5 Large due to the aggressive step compression of its base model; nevertheless, its Vendi (2.259) and LPIPS (0.596) both exceed FLUX, and its GPT-4o scores are the highest overall.

Third, in the GPT-4o-based creative assessment, our SCDiff excels across all three dimensions—novelty, surprise, and value—achieving the highest scores. This demonstrates its capacity to expand conceptual semantic boundaries and effectively access the long-tail distribution regions, resulting in more creative generations.

% \textbf{Quantitative Results}
% Figs. \ref{fig:creative-comparable-result} and \ref{fig:adjectives-comparable-result} show visual comparisons on \textit{CreaCOF} and \textit{AdjCOF}, respectively. We can see that our SCDiff  delivers creative, meaningful and impressive results, outperforming the related SOTA methods. Specifically, the C3 method produces noticeable visual artifacts under high amplification factors, which substantially compromises both output fidelity and creative expression. ConceptLab's reliance on sub-category extraction from broad parent classes leads to inaccurate semantic alignment with target descriptors and significant deviation from intended creative concepts. While GPT-4o maintains content coherence, its outputs fail to achieve the necessary distributional shift required for meaningful creative generation, thereby constraining its imaginative potential.

\textbf{Qualitative Results.}
Figs. \ref{fig:creative-comparable-result} and \ref{fig:adjectives-comparable-result} present visual comparisons on both the CreaCOF and AdjCOF benchmarks, respectively. The results demonstrate that our method, SCDiff, generates creative, meaningful, and visually impressive outputs, surpassing related state-of-the-art approaches. Specifically, C3 exhibits noticeable visual artifacts under high amplification factors, substantially compromising both fidelity and creative expression. ConceptLab relies on sub‑category extraction from broad parent classes, which often leads to inaccurate semantic alignment with target descriptors and a significant deviation from the intended creative concepts. Meanwhile, although GPT-4o, SD 3.5 Large, and Flux.1 Dev maintain high content coherence, their outputs lack the necessary distributional shift required for meaningful creative generation, thereby limiting their imaginative potential.

\begin{figure}[t]
    \centering
    \begin{minipage}[t]{0.48\textwidth}
        \centering
        \includegraphics[width=\linewidth]{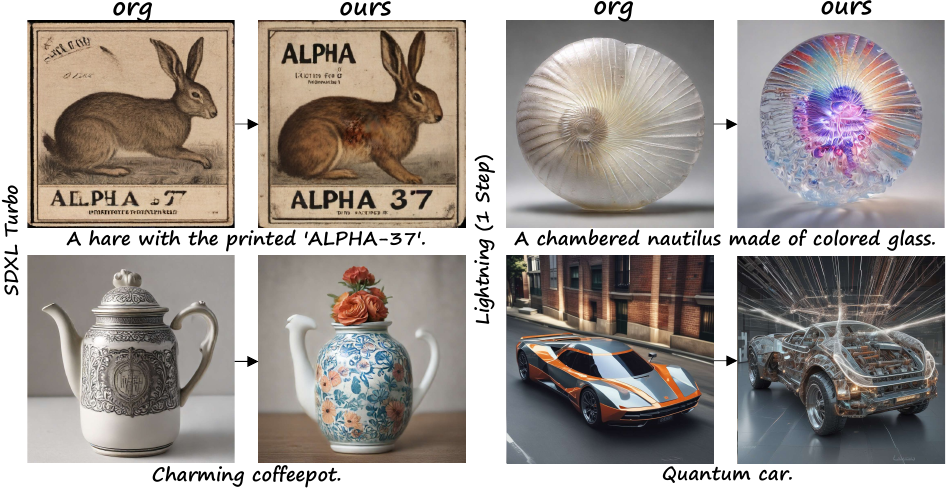}
        \caption{Qualitative results of complex textual inputs using different baseline models.}
        \label{fig:complex-result}
    \end{minipage}
    \hfill
    \begin{minipage}[t]{0.48\textwidth}
        \centering
        \includegraphics[width=\linewidth]{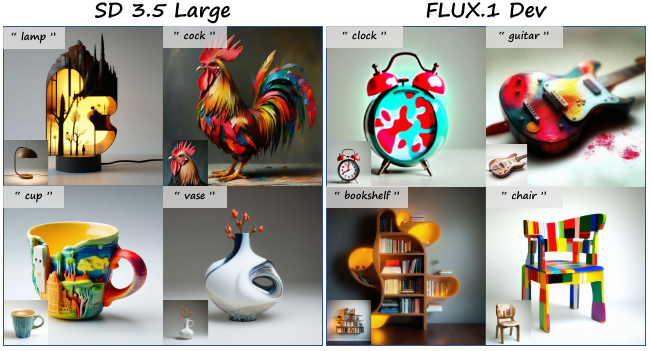}
        \caption{Our method on DiT-based models: (a) SD 3.5, (b) FLUX.}
        \label{fig:dit_results}
    \end{minipage}
    
\vspace{-5mm}
\end{figure}

% \setlength{\abovecaptionskip}{1pt}
% \setlength{\belowcaptionskip}{1pt} 
% \begin{figure}[h]
% \centering
% %\framebox[4.0in]{$\;$}
% \includegraphics[width=0.47\textwidth]{images/15_beta_amblation.pdf}
% \caption{Visual comparisons of the global amplification factors $\alpha$ across different fixed spatial parameters. }
% \label{fig:beta_amblation}
% \vskip -0.1in
% \end{figure}

% \setlength{\columnsep}{8pt}
% \begin{wrapfigure}[17]{r}{0.5\textwidth}
% % \begin{wrapfigure}{r}{0.5\textwidth}
%   \begin{center}
%   \vspace{-15pt}
%     \includegraphics[width=0.48\textwidth]{images/6_complex result.pdf}
%   \vspace{-10pt} 
%   \end{center}
%   \caption{Qualitative results of complex textual inputs using different baseline generative models.}
%   \label{fig:complex-result}
% \end{wrapfigure}

In addition, we explored complex textual prompts with contextual expressions. As shown in Fig. \ref{fig:complex-result}, these prompts improve image detail and semantic consistency without sacrificing realism, demonstrating strong potential for semantically grounded generation. More results are provided in \textbf{Appendix} \ref{sec:more_results}. Specifically, we provide a visualization of the effects of feature amplification across different network layers in Figs. \ref{fig:moreresult1}-\ref{fig:moreresult5}, and demonstrate the generative diversity of fixed prompts under varying random seeds in Figs. \ref{fig:moreresult6}-\ref{fig:moreresult7}.

\vspace{-3mm}
\subsection{Extension to Diffusion Transformers}
\label{sec:dit}
We evaluate our method on Transformer-based diffusion backbones, with qualitative results shown in Fig.~\ref{fig:dit_results}. Specifically, for SD 3.5~\cite{sd3.5} and FLUX~\cite{flux}, latent tokens are first reshaped from sequence format to 2D spatial feature maps for masking, then restored to sequence format for subsequent transformer layers. Despite the absence of skip-connections and hierarchical features in DiT blocks, our method preserves semantic consistency while enhancing diversity, verifying that the core frequency modulation principle is architecture-agnostic.
% between our method and four existing methods.
% Figure \ref{fig:creative-comparable-result} showcases the results with the "creative" prefix, and Figure \ref{fig:adjectives-comparable-result} displays those paired with uncommon out-of-distribution(OOD) adjectives. Qualitative analysis reveals distinct limitations in baseline approaches. 

% In addition to the structured prompts mentioned above, we also explored complex textual prompts that include contextual expressions. As shown in Figure \ref{fig:complex-result}, we enhanced the details and semantic consistency while maintaining the realism of the image, demonstrating its potential in semantically grounded text-to-image generation.

% \begin{figure}[ht]
% \centering
% %\framebox[4.0in]{$\;$}
% \includegraphics[width=0.47\textwidth]{images/6_complex result.pdf}
% \caption{Qualitative results of complex textual inputs using different baseline generative models.}
% \label{fig:complex-result}
% \vskip -0.1in
% \end{figure}

% \setlength{\abovecaptionskip}{1pt}
% \setlength{\belowcaptionskip}{1pt} 
\begin{figure}[t]
    \centering
    \begin{minipage}[t]{0.58\textwidth}
        \centering
        \includegraphics[width=\linewidth]{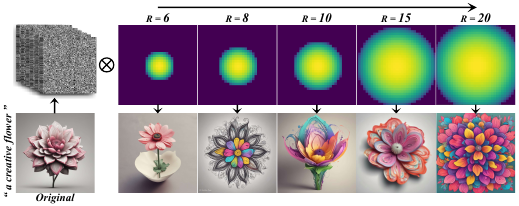}
        \caption{Visual comparisons of different sampling radii ($R$) under the optimal magnification factor $\alpha=3.1$ and shape parameter $\beta=6.7$.}
        \label{fig:R_ablation}
    \end{minipage}
    \hfill
    \begin{minipage}[t]{0.4\textwidth}
        \centering
        \includegraphics[width=\linewidth]{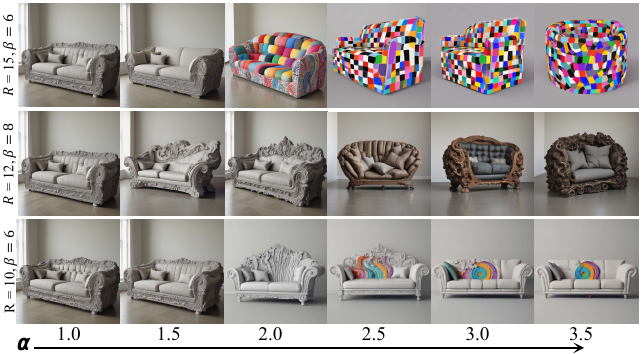}
        \caption{Visual comparisons of the global amplification factors $\alpha$ across different fixed spatial parameters. }
        \label{fig:beta_amblation}
    \end{minipage}
\vspace{-3mm}
\end{figure}

% \begin{figure}[t]
%     \centering
%     \includegraphics[width=0.5\textwidth]{images/10_R_ablation.pdf}
%     \caption{figure}{Visual comparisons of different sampling radii ($R$) under the optimal magnification factor $\alpha=3.1$ and shape parameter $\beta=6.7$.} 
%     \label{fig:R_ablation}
% \end{figure}

\begin{table*}[!htb]
\centering
\setlength{\tabcolsep}{3pt} 
\renewcommand{\arraystretch}{1.2} 
\resizebox{0.97\linewidth}{!}{
\begin{tabular}{l|ccc|ccccc|cccc}
\Xhline{1.2pt}

\multirow{2}{*}{Models} & \multicolumn{3}{c|}{Semantic Alignment} & \multicolumn{5}{c|}{Creativity Metrics} & \multicolumn{4}{c}{Creativity Evaluated by GPT-4o} \\
\cline{2-13}
 & CLIP$\uparrow$ & LLaVA$\uparrow$ & P-tile$\downarrow$ & FID*$\uparrow$ & Precision$\downarrow$ & Recall$\uparrow$ & LPIPS$\uparrow$ & Vendi $\uparrow$ & Surprise$\uparrow$ & Value$\uparrow$ & Novelty$\uparrow$ & Overall$\uparrow$\\
\hline
Baseline w/o LSW and VSML & 0.290 & 0.742 & 0.025 & 664.717 & 0.997 & 0.500 & 0.520 & 1.737& 5.4 & 6.0 & 5.8& 6.0 \\

w/ LSW w/ Fixed $\alpha$    & 0.291 & 0.815 & 0.023 & 705.833 & 0.942 & 0.615 & 0.548 & 1.982 & 6.8 & 6.1 & 6.1 & 6.4 \\

w/ LSW w/ Fixed $\beta$   & 0.301 & 0.834 & 0.021 & 692.415  & 0.981 & 0.554 & 0.539 & 1.645 & 6.3 & 6.9 & 6.8 & 6.9 \\

w/ LSW and VSML (\textbf{our Full Model}) & \textbf{0.308} & \textbf{0.911} & \textbf{0.024} & \textbf{722.057} & \textbf{0.958} & \textbf{0.719} & \textbf{0.556} & \textbf{2.188} & \textbf{7.8} & \textbf{7.2} & \textbf{7.8} & \textbf{7.8} \\
\Xhline{1.2pt}
\end{tabular}}
\caption{Quantitative ablation study of our proposed SCDiff.}
\label{tab:ablation_analysis}
\vskip -0.2in
\end{table*}

\setlength{\columnsep}{8pt}
\begin{wrapfigure}[5]{r}{0.45\textwidth}
% \begin{wrapfigure}{r}{0.5\textwidth}
  \begin{center}
  \vspace{-30pt}
    \includegraphics[width=0.45\textwidth]{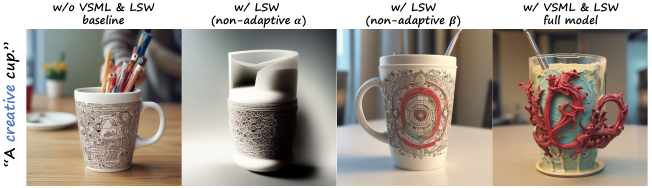}
  \vspace{-30pt} 
  \end{center}
\caption{Ablation analysis.}
\label{fig:ablation}
\end{wrapfigure}
\vspace{-5mm}
\subsection{Ablation Study}
We conducted an ablation study to evaluate the contributions of each key component in SCDiff, as shown in Fig.~\ref{fig:ablation} and Table~\ref{tab:ablation_analysis}. For the configuration with $R=15$, the experiments proceeded through the following progressive modifications, shown from left to right: (i) baseline model without the LSW and VSML modules; (ii) With the LSW module (non adaptive amplification factor $\alpha$);
(iii) With the LSW module (non adaptive Kaiser–Bessel window ($\beta$));
(iv) Full model with both LSW and VSML modules. Experimental results confirm that the adaptive design of each component meaningfully improves both image quality and creative effectiveness.

%We progressively remove or modify core modules: (i) baseline model (removing both VSML and LSW modules); (ii) replacing center-focused enhancement with edge-focused enhancement; (iii) replacing the Kaiser–Bessel window with a Gaussian window; (iv) replacing the Kaiser–Bessel window with a binary circular mask (i.e., a flat window with unit response inside $r \le R$); (v) using a frequency-domain variant based on the zeroth-order modified Bessel function; (vi) fixed $\alpha$ search mechanism in VSML module ($\alpha=5$), we ablate only $\alpha$ here, as $\beta$ and R are continuous parameters that are better examined in the parameter analysis section \ref{sec:param_analysis}. Experimental results demonstrate that the adaptive design of each component significantly enhances both image quality and creative effectiveness.

\vspace{-3mm}
\subsection{Parameter Analysis}
\label{sec:param_analysis}
\textbf{Sampling radius $R$.} As illustrated in Fig. \ref{fig:R_ablation}, we visualize the impact of varying the sampling radius $R$ on the \textit{``flower''} concept, while fixing the global amplification factor $\alpha=3.1$ and the shape parameter $\beta=6.7$ (derived from our optimization process). Specifically, a large radius causes the flower to occupy the entire canvas, whereas a small radius restricts the deformation to only the central circular region. This observation confirms that $R$ effectively serves as a flexible control knob, allowing users to customize the spatial extent of creative editing according to their design intent, thereby producing diverse visual outcomes. 
% While $R=10$ provides a more localized fit for this specific object, $R=15$ is adopted as default for its superior robustness across various object scales.

%\textbf{Parameter Analysis.} As illustrated in Figure \ref{fig:R_ablation}, we visualize the impact of varying the sampling radius $R$ on the ``flower'' concept, while fixing the global amplification factor $\alpha=3.1$ and the shape parameter $\beta=6.7$ (both derived from our optimization process). Specifically, a large radius causes the flower to occupy the entire canvas, whereas a small radius restricts the deformation to only the central circular region. This observation confirms that $R$ effectively serves as a flexible control knob, allowing users to customize the spatial extent of creative editing according to their design intent, thereby producing diverse visual outcomes.

\textbf{Amplification factor $\alpha$.} Can optimizing the global amplification factor $\alpha$ compensate for a fixed but suboptimal spatial window? As illustrated in Fig. \ref{fig:beta_amblation}, fixing the spatial parameters ($R$, $\beta$) and increasing $\alpha$ alone reveals limitations. If the fixed window is too small for the concept, edits remain strictly localized even with large $\alpha$, preventing the global structural reconstruction. Conversely, an oversized window that intrudes into the background leads to structural degradation as $\alpha$ increases, driving identity similarity $\mathcal{S}_{\text{img}}$ below the safety threshold $\tau$. Thus, our two-stage strategy enables multi-scale feature manipulation, enhancing visual novelty while maintaining structural integrity.

\vspace{-3mm}
\section{Conclusion}
\vspace{-3mm}
In this paper, we proposed SCDiff, an adaptive and semantic-aware framework for creative object generation in text-to-image synthesis. We incorporate the U-Net architecture by inserting a Learnable Spatial Weighting (LSW) module into the shallow encoder layers, where a parametric Kaiser–Bessel window selectively amplifies core structural features and fosters the emergence of novel visual patterns. In parallel, we introduce the Visual-Semantic Mixing Loss (VSML) module, which employs a dual-objective loss to jointly encourage semantic alignment and controlled visual divergence, enabling more expressive and creative generations. Extensive experiments across 950 object categories validate that SCDiff significantly outperforms existing methods in terms of visual creativity, semantic faithfulness, and structural coherence. This establishes a new direction for post-alignment creative generation and delivers an efficient tool for digital art and game design.

\bibliographystyle{IEEEtran}
\bibliography{neurips_2026}

% \begin{ack}
% Use unnumbered first level headings for the acknowledgments. All acknowledgments

% Do {\bf not} include this section in the anonymized submission, only in the final paper. You can use the \texttt{ack} environment provided in the style file to automatically hide this section in the anonymized submission.
% \end{ack}

%%%%%%%%%%%%%%%%%%%%%%%%%%%%%%%%%%%%%%%%%%%%%%%%%%%%%%%%%%%%
\newpage
\appendix

This supplementary material provides additional technical details and extended results to support the main paper. First, in Section \ref{sec:feature_enhancement_discussion}, we analyze the impact of feature enhancement across different U-Net blocks. Section \ref{sec:Hierarchical} details the hierarchical Bayesian-SPSA optimization procedure to solve the constrained optimization problem. In Section \ref{sec:limitation}, we discuss the limitations of our method. Section \ref{sec:datasets} details the complete taxonomy of the COF dataset, along with the strategy for assigning tailored adjectival prompts designed for complex generation tasks. Section \ref{sec:userstudy} provides specific details of the user study and a detailed breakdown of the final vote counts for each question. Finally, in Section \ref{sec:more_results}, we showcase additional qualitative results. Section \ref{sec:impact} discusses the broader societal impacts of our method, including both positive applications and potential misuse risks.

\setlength{\abovecaptionskip}{1pt}
\setlength{\belowcaptionskip}{1pt} 
\begin{figure*}[h]
\begin{center}
%\framebox[4.0in]{$\;$}
\includegraphics[width=0.97\textwidth]{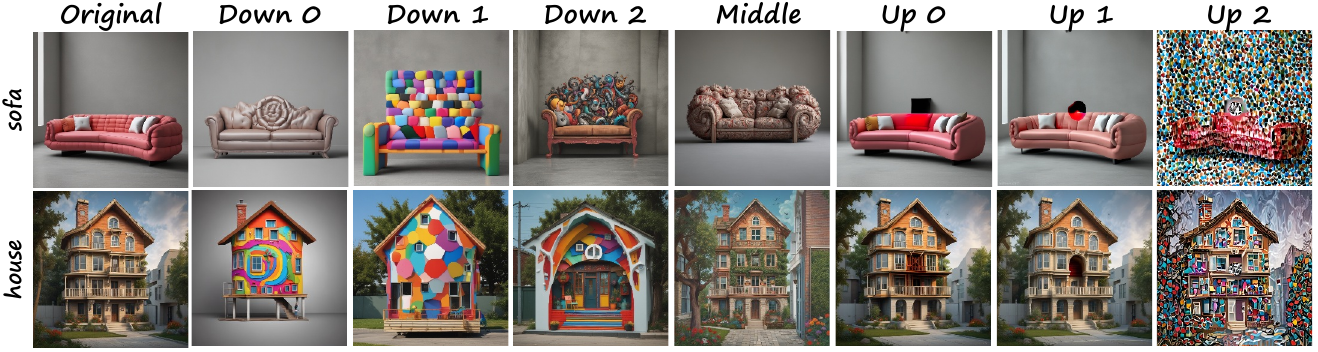}
\end{center}
\caption{The images generated using the adaptively optimized parameters for each block.}
\vspace{-10pt} 
\label{fig:block_analysis}
\end{figure*}
\vspace{-3mm}

\section{Discussion on Feature Enhancement: Shallow vs. Deep.}
\label{sec:feature_enhancement_discussion}
As shown in Figure \ref{fig:block_analysis}, experiments demonstrate that enhancing downsampling and middle blocks yields the best gains in structural creativity. Applying the module in upsampling layers adds noise because these layers focus on texture refinement rather than altering spatial structure. Therefore, we apply feature enhancement only to downsampling and middle blocks.

\section{Hierarchical Optimization}
\label{sec:Hierarchical}

\subsection{Bayesian Optimization for Amplification Factor $\alpha$}
\label{sec:bayesopt}
\paragraph{Surrogate model.} We employ a Gaussian Process (GP) as the surrogate for the implicit objective $\tilde{f}(\alpha_k)$, using the Mat\'ern-5/2 kernel:
\begin{equation}
k(\alpha, \alpha') = \sigma^2 \left(1 + \frac{\sqrt{5}r}{l} + \frac{5r^2}{3l^2}\right) \exp\left(-\frac{\sqrt{5}r}{l}\right),
\end{equation}
where $r = |\alpha - \alpha'|$. We assume a zero-mean GP prior and independent Gaussian observation noise $\varepsilon \sim \mathcal{N}(0, \sigma_n^2)$. The length scale $l$, signal variance $\sigma^2$, and noise variance $\sigma_n^2$ are jointly inferred by maximizing the marginal likelihood.

\paragraph{Acquisition function.} We adopt Expected Improvement (EI), which under the GP posterior $\tilde{f}(\alpha) \mid \mathcal{D}_n \sim \mathcal{N}(\mu_n(\alpha), \sigma_n^2(\alpha))$ admits the closed-form expression:
\begin{equation}
\text{EI}(\alpha) = (\mu_n(\alpha) - \tilde{f}^+) \Phi(Z) + \sigma_n(\alpha) \phi(Z),
\end{equation}
where $\tilde{f}^+ = \max_{1 \le i \le n} \tilde{f}(\alpha_i)$ is the current best observation, $Z = (\mu_n(\alpha) - \tilde{f}^+) / \sigma_n(\alpha)$, and $\Phi(\cdot), \phi(\cdot)$ are the standard normal CDF and PDF, respectively. EI automatically balances exploitation of promising regions and exploration of high-uncertainty areas.

\paragraph{Optimization procedure.} The search space is $[1.5, 8.0]$. We initialize with $n_{\text{init}} = 5$ random samples via Latin hypercube sampling to ensure space coverage, followed by $n_{\text{iter}} = 10$ sequential acquisition iterations. At each iteration, we maximize EI via L-BFGS-B with 10 random restarts to avoid local optima, query the point with maximal EI, observe $\tilde{f}(\alpha_{n+1})$, and update the GP posterior. The optimal amplification factor is selected as the point with maximum posterior mean: $\alpha_k^* = \arg\max_{\alpha} \mu_n(\alpha)$. With only $n = 15$ total observations, exact GP inference via Cholesky decomposition ($\mathcal{O}(n^3)$) is computationally negligible.

\subsection{SPSA for Constrained Window Parameter $\beta$}
\label{sec:SPSA}
\paragraph{Algorithm.} Simultaneous Perturbation Stochastic Approximation (SPSA)~\cite{spall1998implementation} is a zeroth-order optimization method that estimates gradients via simultaneous random perturbations, requiring only two function evaluations per iteration regardless of parameter dimensionality.

At iteration $t$, we generate a random perturbation vector $\Delta_t$ with entries drawn independently from a symmetric Bernoulli distribution ($\pm 1$ with equal probability). The pseudo-gradient is estimated as:
\begin{equation}
\hat{\nabla} f(\beta_t) = \frac{f(\beta_t + c_t \Delta_t) - f(\beta_t - c_t \Delta_t)}{2 c_t} \odot \Delta_t,
\end{equation}
where $c_t = c / t^\gamma$ is the decaying perturbation size with $c = 0.1$ and $\gamma = 0.101$, and $\odot$ denotes element-wise product (since $1/\Delta_t^{(i)} = \Delta_t^{(i)}$ for $\Delta_t^{(i)} \in \{-1, +1\}$). The parameter is updated via projected gradient ascent:
\begin{equation}
\beta_{t+1} = \Pi_{[\beta_{\min}, \beta_{\max}]}\left(\beta_t + a_t \hat{\nabla} f(\beta_t)\right),
\end{equation}
where $a_t = a / t^\alpha$ is the decaying learning rate with $a = 0.5$, and $\Pi(\cdot)$ denotes projection onto the feasible box.

\paragraph{Constraint handling.} If the updated $\beta_{t+1}$ violates $g(\beta_{t+1}) > 0$ (i.e., insufficient deformation), we halve the step size $a_t \leftarrow a_t / 2$ and resample the perturbation direction $\Delta_t$ until a feasible point is found. This ensures that the optimization trajectory remains within the feasible region while maximizing text alignment.

\paragraph{Implementation details.} We set the initial value $\beta_0 = 8.0$, with search interval $[\beta_{\min}, \beta_{\max}] = [6.0, 12.0]$ and run for $T = 50$ iterations. To mitigate SPSA's inherent stochasticity, we perform 5 independent runs with different random seeds and select the final $\beta_k^*$ as the one achieving the highest feasible objective value.

\setlength{\abovecaptionskip}{1pt}
\setlength{\belowcaptionskip}{1pt} 
\begin{figure*}[h]
\begin{center}
%\framebox[4.0in]{$\;$}
\includegraphics[width=0.6\textwidth]{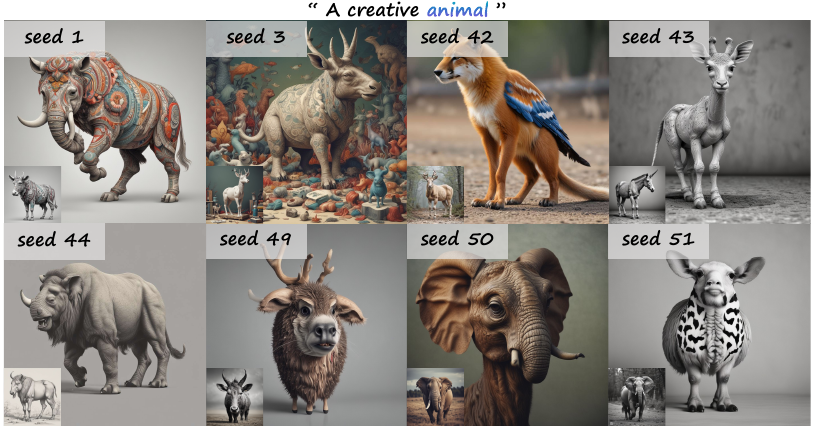}
\end{center}
\caption{Qualitative results for intra-class diversity under the prompt ``a creative animal''.}
\label{fig:animal}
\end{figure*}

% \setlength{\abovecaptionskip}{1pt}
% \setlength{\belowcaptionskip}{1pt} 
% \begin{figure*}[h]
% \begin{center}
% %\framebox[4.0in]{$\;$}
% \includegraphics[width=0.6\textwidth]{images/sd3.5.pdf}
% \end{center}
% \caption{Qualitative results on DiT architectures. (a) SD 3.5, (b) FLUX. Our frequency-domain modulation transfers to transformer backbones without modification, preserving text alignment while enhancing visual diversity.}
% \label{fig:dit_results}
% \end{figure*}

% \section{Future Work and Open Challenges}
% While this work focuses on the dominant central subjects to maximize creative impact, the LSW module's spatial anchor is parameterizable and can be shifted via simple coordinate offsets to accommodate off-center or multi-object compositions if necessary.

\section{Limitation}
\label{sec:limitation}
\subsection{Structural Constraints in Fine-Grained Generation}

Our method operates under the constraint of semantic fidelity, exhibiting limitations when processing fine-grained categories with strong structural priors. As shown in Fig.~\ref{fig:animal}, under broad concepts (e.g., ``a creative animal''), the model generates pronounced diversity in both structure and appearance; however, when the prompt is constrained to a specific species, the model attempts structural variations yet these tend to converge to similar silhouette patterns, rendering structural differences across samples barely perceptible to human observers, with innovations primarily reflected in texture pattern variations. To alleviate this tension, future work could introduce explicit cross-category blending mechanisms, enabling more directed creative exploration while maintaining semantic consistency.

% Our approach exhibits limitations when processing categories with homogeneous visual characteristics. As shown in Fig.~\ref{fig:animal}, under a broad concept (e.g., ``a creative animal''), the model is able to generate structures and appearances with pronounced diversity; however, when the prompt is constrained to a specific animal species, the model’s innovations are mainly reflected in variations in texture patterns, with limited ability to produce substantial structural contour reconfiguration.
% This constraint arises from inadequate geometric diversity within the training dataset. To address this limitation, future work could incorporate explicit cross-category blending mechanisms. This would transform the innovation process from being predominantly texture-driven to one that encompasses controlled structural reconfiguration, enabling more directed creative exploration.

\setlength{\abovecaptionskip}{1pt}
\setlength{\belowcaptionskip}{1pt} 
\begin{figure*}[h]
\begin{center}
%\framebox[4.0in]{$\;$}
\includegraphics[width=0.6\textwidth]{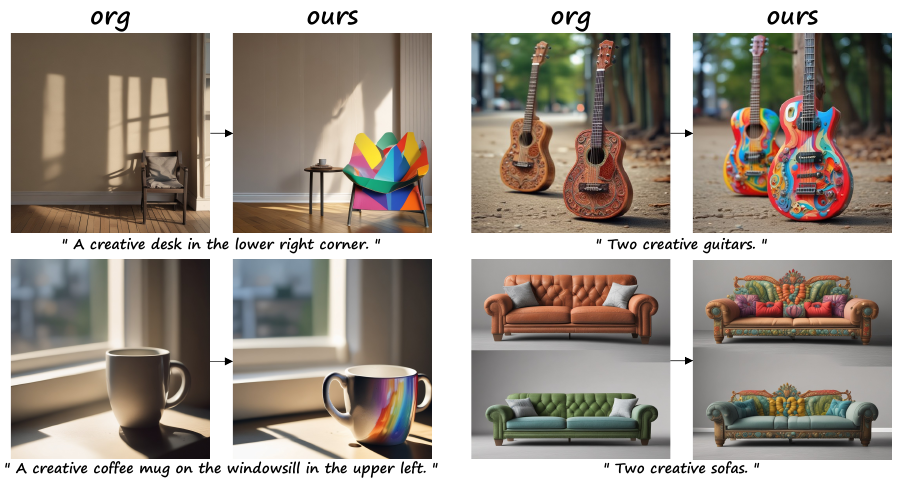}
\end{center}
\caption{Qualitative results for intra-class diversity under the prompt ``a creative animal''.}
\label{fig:notcenter}
\end{figure*}

\subsection{Spatial Prior Assumptions}
To accommodate the inherent central-cropping bias of dominant T2I backbones, \textbf{SCDiff} defaults to a center-aligned window, with quantitative evaluation focused on single-object creativity. The spatial anchor $(x_c, y_c)$ is user-adjustable and supports off-center object enhancement. Fig.~\ref{fig:notcenter} (left) shows precise enhancement for corner-placed objects; (right) shows coherence in multi-object scenes under moderate amplification factors. Generalizing this localized control to complex multi-subject environments remains future work, which could be facilitated by integrating autonomous object-localization priors to drive the LSW module.

\section{Datasets}
\label{sec:datasets}

We adopt the COF dataset introduced by AGSwap \cite{AGSwapOC} as our evaluation benchmark. This dataset effectively reorganizes ImageNet-1k categories based on the WordNet hierarchy and manual curation to ensure visual similarity. It comprises 95 visually consistent superclasses (e.g., retaining 'dog' while excluding 'wild dog') with 10 balanced subclasses each. To rigorously assess the model's generative novelty across varying difficulty levels, we derive two distinct benchmarks from this structure: a Simple Dataset, constructed by appending the prefix ``creative'' to each of the 950 subclasses to evaluate basic diversity; and a Complex Dataset, which assigns 10 tailored adjectives to each superclass, yielding 9,500 prompt variants to challenge the model with higher structural complexity. 
% The comprehensive list of all 950 categories and their corresponding adjectives for the complex set is provided in Tables \ref{tab:Detail_adj1}--\ref{tab:Detail_adj6}.

\section{User Study}
% To evaluate the creativity of the generated images based on human perception, we conducted a user study (Fig. \ref{fig:userstudyyw}). A total of 187 participants rated 10 output sets (5 for creative generation and 5 for OOD adjectives), yielding 1870 valid ratings. Our proposed SCDiff method achieved the highest preference scores in both task categories: 81.7\% for the creative generation task and 80.3\% for the mixing/editing task, significantly outperforming the original model (9.7\%, 6.1\%), C3 (1.9\%, 3.2\%), and GPT-4o (4.4\%, 8.0\%). These results confirm that SCDiff aligns more closely with human aesthetic preferences, excelling in both visual coherence and creativity. 
% Complete experimental details are available in \textbf{Appendix} \ref{sec:userstudy}.

\label{sec:userstudy}
\setlength{\abovecaptionskip}{1pt}
\setlength{\belowcaptionskip}{1pt} 
\begin{figure}[h]
\begin{center}
%\framebox[4.0in]{$\;$}
\includegraphics[width=0.5\textwidth]{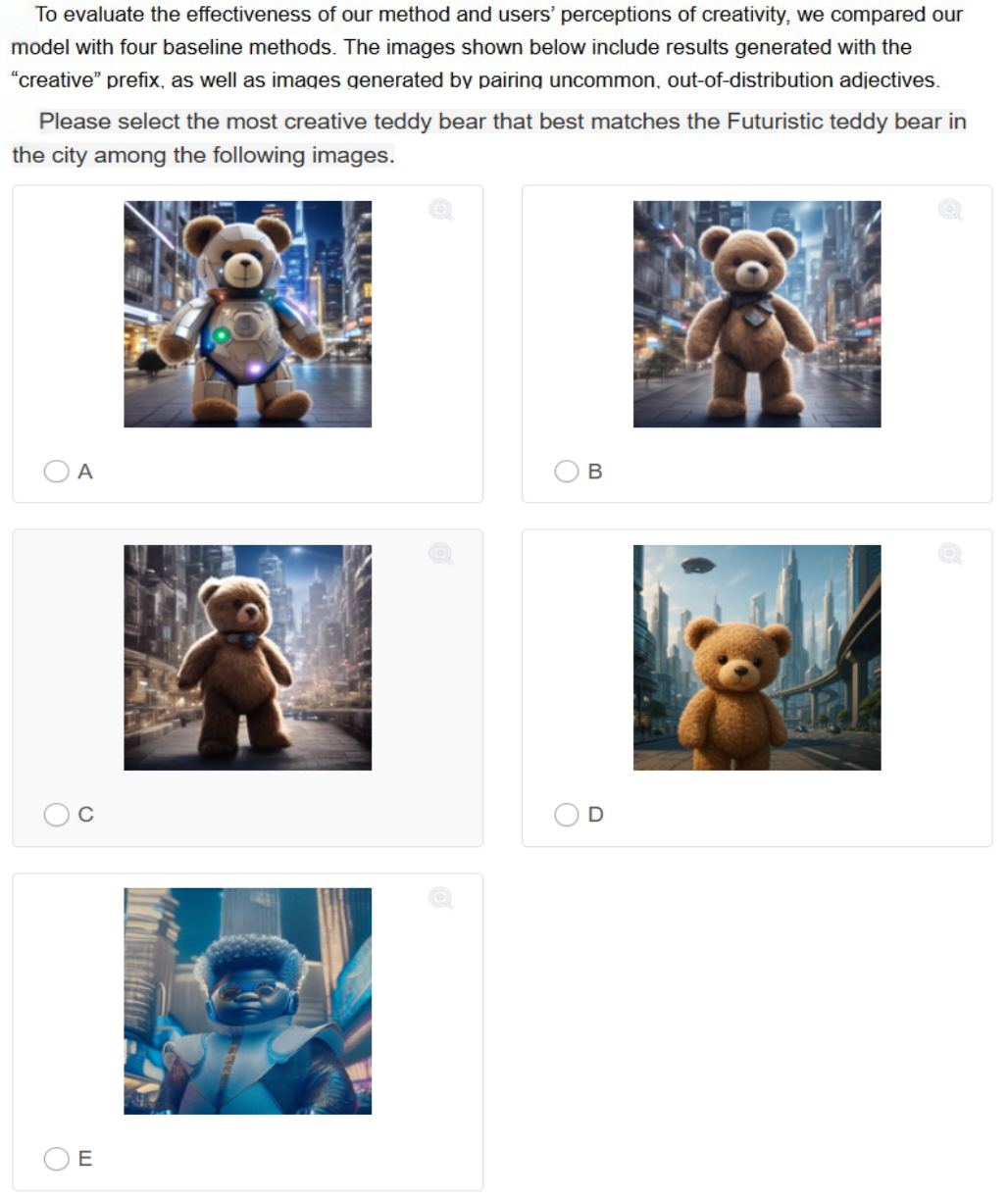}
\end{center}
\caption{The user interface used for the human evaluation task. Participants were instructed to choose based on visual quality and alignment with the text prompt.}
\label{fig:userstudy}
\end{figure}

\begin{figure}[t]
\begin{center}
%\framebox[4.0in]{$\;$}
\includegraphics[width=0.47\textwidth]{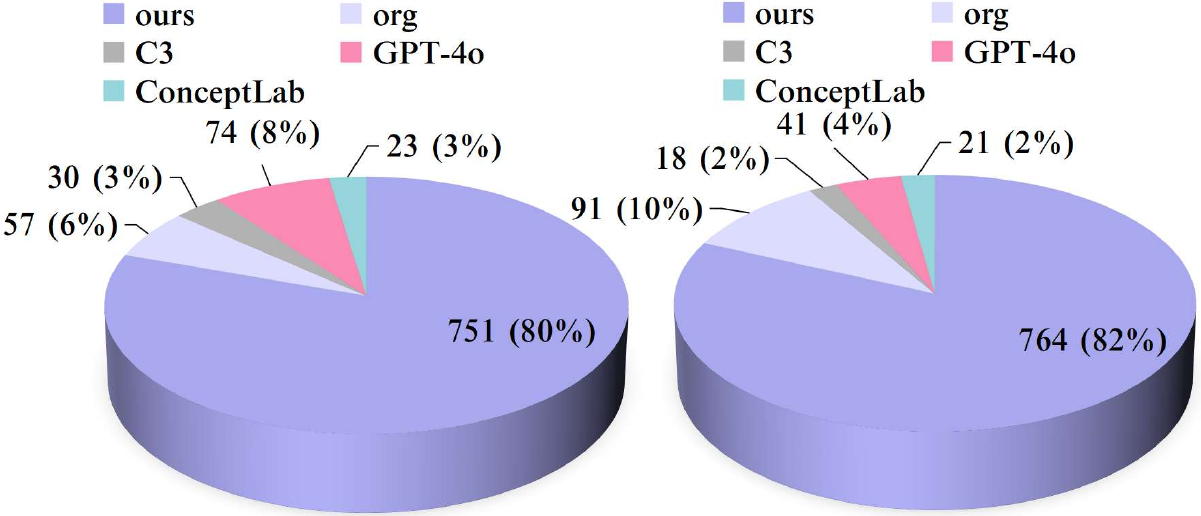}
\end{center}
\caption{User study comparing different methods on creative prompts (left, CreaCOF) and adjective prompts (right, AdjCOF).}
\label{fig:userstudyyw}
\vskip -0.1in
\end{figure}

% \setlength{\abovecaptionskip}{1pt}
% \setlength{\belowcaptionskip}{1pt} 
% \begin{figure}[h]
% \begin{center}
% %\framebox[4.0in]{$\;$}
% \includegraphics[width=0.5\textwidth]{images/12_userstudy.pdf}
% \end{center}
% \caption{User studies.}
% \label{fig:userstudyyw}
% \end{figure}

To evaluate the effectiveness of our method and users' perceptions of creativity, we conducted a comprehensive user study comparing our method against state-of-the-art baselines: C3, Org, GPT-4o, and ConceptLab. The evaluation focused on two distinct categories: OOD Adjectives, which involve complex attribute bindings and material conflicts (e.g., ``Frozen flame vase'', ``Rock paper boat''), and Creative Generation, which involves open-ended prompts (e.g., ``Creative bed''). The detailed vote distributions for these categories are presented in Table~\ref{tab:user_study_combinational_ood} and Table~\ref{tab:user_study_combinational_creative}, respectively. Figure \ref{fig:userstudyyw} further illustrates the percentage distribution of votes across all compared methods. Figure~\ref{fig:userstudy} presents examples of the user study questionnaires. A total of 193 participants completed the survey, with each participant evaluating 10 fusion results (5 from each group), contributing a total of 1,930 votes. Participants were instructed to select the result that demonstrated the highest creativity and best alignment with the text prompt.

As evidenced by the quantitative results, participants showed a strong preference for our method. In the OOD Adjectives category (Table~\ref{tab:user_study_combinational_ood}), our method demonstrated exceptional capability in resolving conflicting semantic attributes. For instance, in the ``Futuristic car'' case, our method secured a dominant \textbf{85.49\%} of the votes, significantly outperforming the second-best method, C3 (8.29\%). Similarly, for abstract material combinations such as ``Frozen flame vase'' and ``Rock paper boat,'' our method garnered \textbf{80.31\%} and \textbf{82.38\%} of the preferences, respectively, attesting to its robustness in synthesizing coherent structures from contradictory textual descriptions. Although C3 occasionally ranked second (e.g., 17.62\% for ``Mountain architecture''), it remained substantially lower than our approach (73.58\%).

In the Creative Generation category (Table~\ref{tab:user_study_combinational_creative}), our method maintained a strong lead. Specifically, for prompts requiring significant structural deviation such as ``Creative mirror'' and ``Creative table,'' our approach garnered an overwhelming 85.49\% and 82.38\% of user votes, respectively. Even in categories with broad design spaces like ``Creative bed'', where GPT-4o showed improved competence (17.10\%), our method maintained a dominant lead with 67.88\%. Results show our framework aligns better with human aesthetic preferences and creative expectations than existing state-of-the-art methods.

\begin{table*}[!t]
\centering
\footnotesize
\caption{User study results on \textbf{Creative Generation}. This category assesses the visual novelty and imagination of objects generated from open-ended ``Creative [class]'' prompts.}
\label{tab:user_study_combinational_ood}
\begin{tabular}{l|c|c|c|c|c}
\hline
\diagbox{prompt}{Method} & A(Ours) & B(C3) & C(Org) & D(GPT-4o) & E(ConceptLab) \\
\hline
Creative table      & 159(82.38\%)& 9(4.66\%)& 5(2.59\%)& 19(9.84\%)& 1(0.52\%)\\
Creative cactus     & 156(80.83\%)& 11(5.7\%)& 9(4.66\%)& 11(5.7\%)& 6(3.11\%)\\
Creative bed        & 131(67.88\%)& 9(4.66\%)& 12(6.22\%)& 33(17.1\%)& 8(4.15\%)\\
Creative mirror     & 165(85.49\%)& 12(6.22\%)& 2(1.04\%)& 10(5.18\%)& 4(2.07\%)\\
\hline
\end{tabular}
\end{table*}

\begin{table*}[!t]
\centering
\footnotesize
\caption{User study results on \textbf{OOD Adjectives}. This category evaluates the model's capability in handling complex attribute bindings and material conflicts.}
\label{tab:user_study_combinational_creative}
\begin{tabular}{l|c|c|c|c|c}
\hline
\diagbox{prompt}{Method} & A(Ours) & B(C3) & C(Org) & D(GPT-4o) & E(ConceptLab) \\
\hline
Frozen flame vase     & 155(80.31\%) & 29(15.03\%)   & 5(2.59\%) & 4(2.07\%) & 0(0\%) \\
Futuristic teddy bear in the city         & 163(84.46\%)  & 12(6.22\%)   & 1(0.52\%)   & 10(5.18\%) & 7(3.63\%) \\
Iridescent storm clouds   & 159(82.38\%) & 19(9.84\%)   & 3(1.55\%) & 4(2.07\%) & 8(4.15\%)\\
Mountain architecture      & 142(73.58\%)& 34(17.62\%)& 2(1.04\%)& 12(6.22\%)   & 3(1.55\%) \\
Rock paper boat  & 159(82.38\%) & 10(5.18\%)& 8(4.15\%)& 7(3.63\%) & 9(4.66\%)\\
Futuristic car       & 165(85.49\%)& 16(8.29\%)& 3(1.55\%)& 8(4.15\%)& 1(0.52\%)\\
\hline
\end{tabular}
\end{table*}

\setlength{\abovecaptionskip}{1pt}
\setlength{\belowcaptionskip}{1pt} 
\begin{figure*}[h]
\begin{center}
%\framebox[4.0in]{$\;$}
\includegraphics[width=0.88\textwidth]{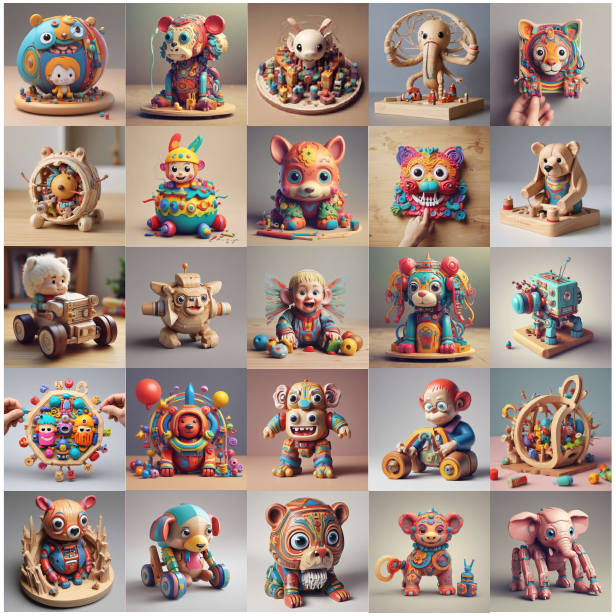}
\end{center}
\caption{More visual results. Additional samples generated using the prompt "a creative toy" are presented.}
\label{fig:moreresult6}
\end{figure*}

\setlength{\abovecaptionskip}{1pt}
\setlength{\belowcaptionskip}{1pt} 
\begin{figure*}[h]
\begin{center}
%\framebox[4.0in]{$\;$}
\includegraphics[width=0.9\textwidth]{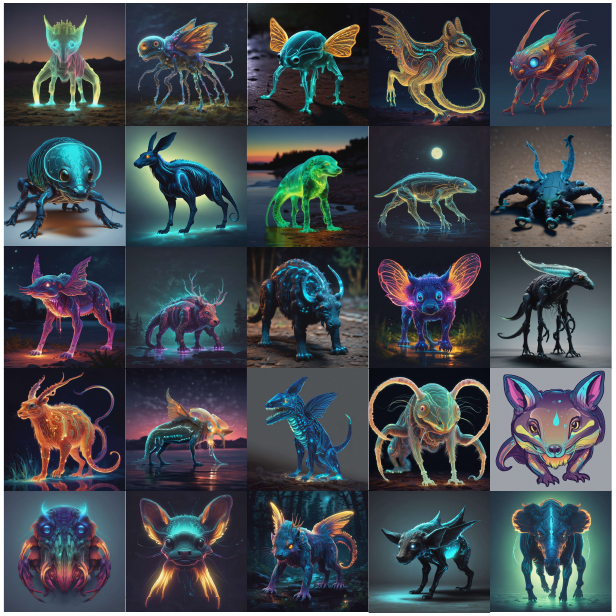}
\end{center}
\caption{More visual results. Additional samples generated using the prompt "a bioluminescent animal" are presented.}
\label{fig:moreresult7}
\end{figure*}

\setlength{\abovecaptionskip}{1pt}
\setlength{\belowcaptionskip}{1pt} 
\begin{figure*}[h]
\begin{center}
%\framebox[4.0in]{$\;$}
\includegraphics[width=0.9\textwidth]{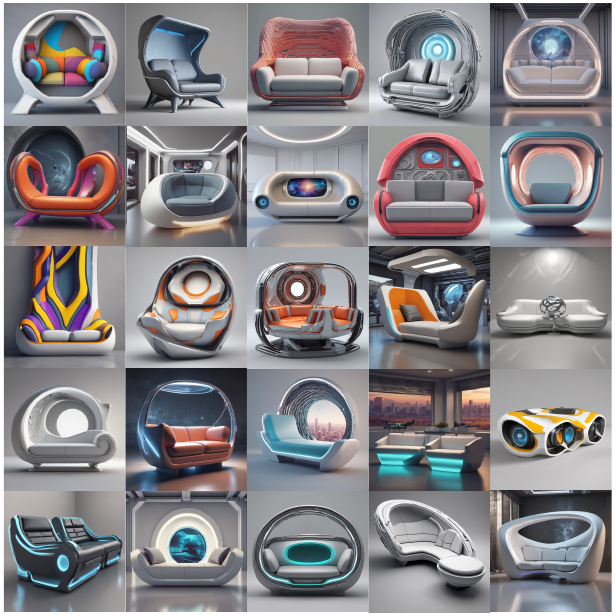}
\end{center}
\caption{More visual results. Additional samples generated using the prompt "a futuristic sofa" are presented.}
\label{fig:moreresult8}
\end{figure*}

\setlength{\abovecaptionskip}{1pt}
\setlength{\belowcaptionskip}{1pt} 
\begin{figure*}[h]
\begin{center}
%\framebox[4.0in]{$\;$}
\includegraphics[width=0.9\textwidth]{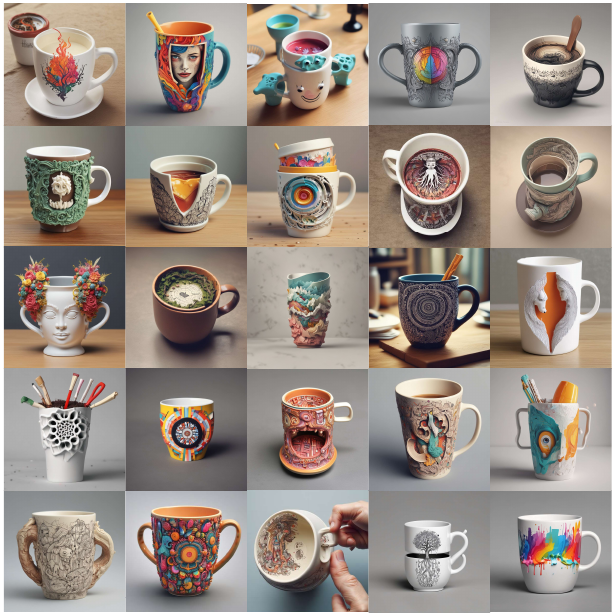}
\end{center}
\caption{More visual results. Additional samples generated using the prompt "a creative cup" are presented.}
\label{fig:moreresult9}
\end{figure*}

\setlength{\abovecaptionskip}{1pt}
\setlength{\belowcaptionskip}{1pt} 
\begin{figure*}[h]
\begin{center}
%\framebox[4.0in]{$\;$}
\includegraphics[width=0.9\textwidth]{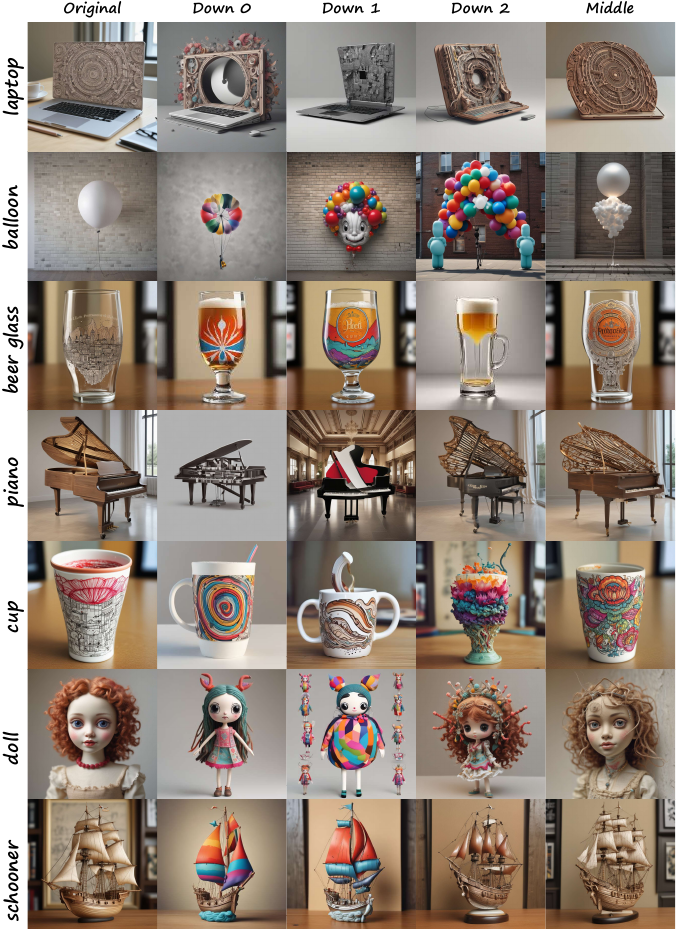}
\end{center}
\caption{More visual results. Visualization of feature amplification across different U-Net layers, confirming the model's versatility in generating novel content at every structural level.}
\label{fig:moreresult1}
\end{figure*}

\setlength{\abovecaptionskip}{1pt}
\setlength{\belowcaptionskip}{1pt} 
\begin{figure*}[h]
\begin{center}
%\framebox[4.0in]{$\;$}
\includegraphics[width=0.9\textwidth]{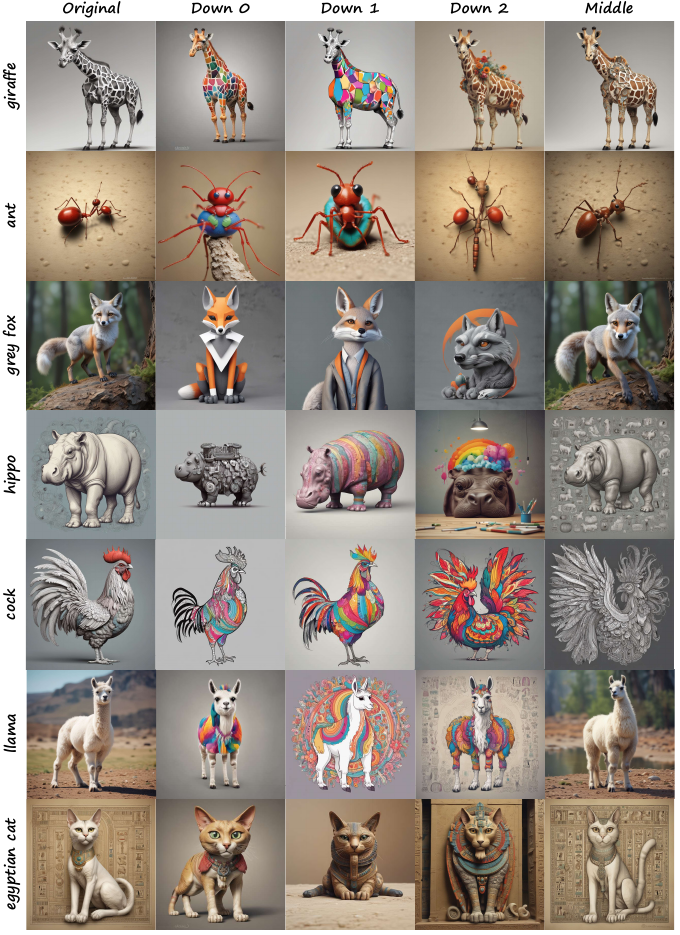}
\end{center}
\caption{More visual results. Visualization of feature amplification across different U-Net layers, confirming the model's versatility in generating novel content at every structural level.}
\label{fig:moreresult2}
\end{figure*}

\setlength{\abovecaptionskip}{1pt}
\setlength{\belowcaptionskip}{1pt} 
\begin{figure*}[h]
\begin{center}
%\framebox[4.0in]{$\;$}
\includegraphics[width=0.9\textwidth]{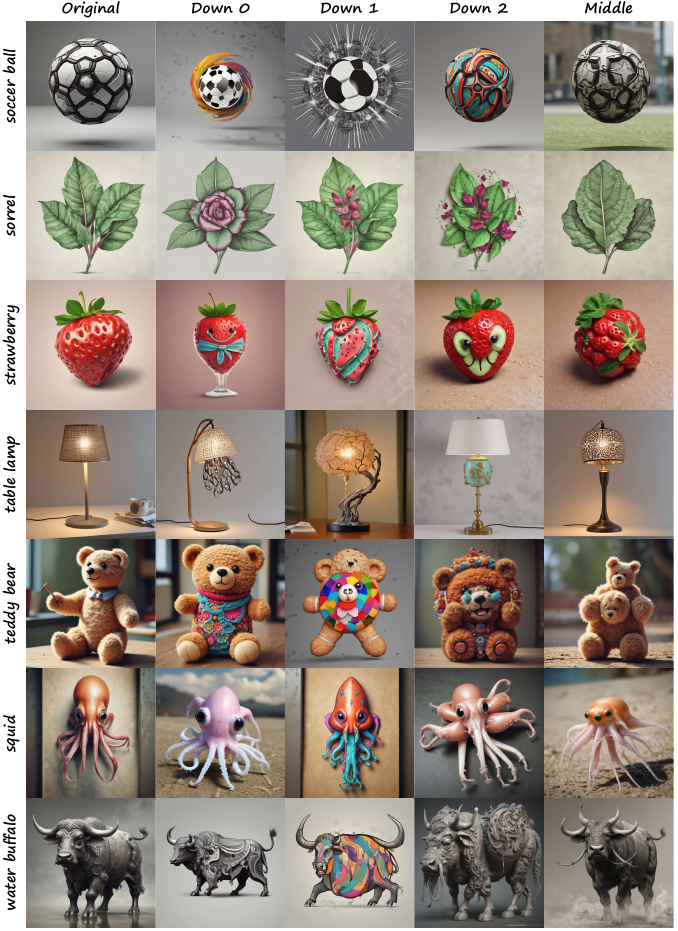}
\end{center}
\caption{More visual results. Visualization of feature amplification across different U-Net layers, confirming the model's versatility in generating novel content at every structural level.}
\label{fig:moreresult3}
\end{figure*}

\setlength{\abovecaptionskip}{1pt}
\setlength{\belowcaptionskip}{1pt} 
\begin{figure*}[h]
\begin{center}
%\framebox[4.0in]{$\;$}
\includegraphics[width=0.9\textwidth]{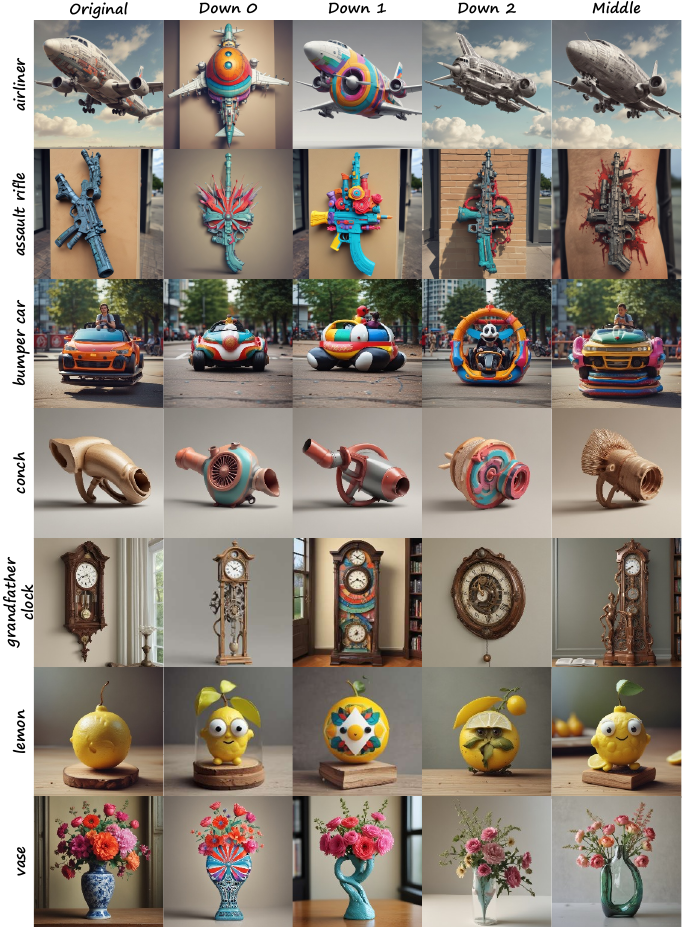}
\end{center}
\caption{More visual results. Visualization of feature amplification across different U-Net layers, confirming the model's versatility in generating novel content at every structural level.}
\label{fig:moreresult4}
\end{figure*}

\setlength{\abovecaptionskip}{1pt}
\setlength{\belowcaptionskip}{1pt} 
\begin{figure*}[h]
\begin{center}
%\framebox[4.0in]{$\;$}
\includegraphics[width=0.9\textwidth]{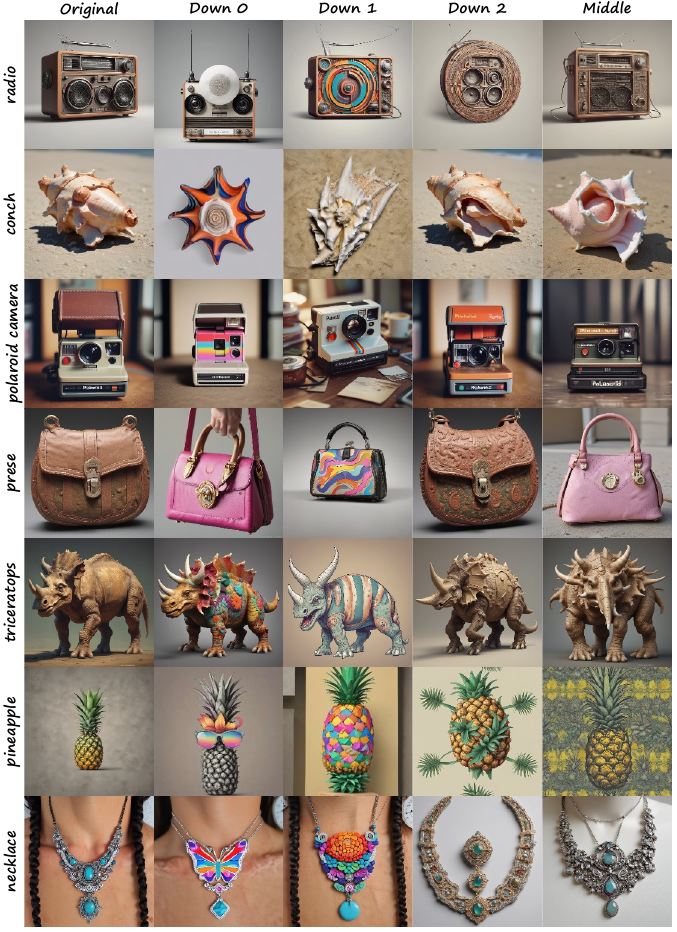}
\end{center}
\caption{More visual results. Visualization of feature amplification across different U-Net layers, confirming the model's versatility in generating novel content at every structural level.}
\label{fig:moreresult5}
\end{figure*}

\section{More Results}
We present an extensive gallery of generated samples to further demonstrate the capabilities and interpretability of our framework.
First, in Fig.~\ref{fig:moreresult6}--\ref{fig:moreresult9}, we showcase the generative diversity of SCDiff. By fixing the textual prompt while varying the random seed, these results illustrate the model's ability to produce a wide range of distinct, creative interpretations for a single concept.
Second, in Figures~\ref{fig:moreresult1}--\ref{fig:moreresult5}, we provide a layer-wise visualization of the feature amplification process. By applying the LSW module to different encoder layers (e.g., Down and Mid blocks) under a fixed seed and prompt, we demonstrate how specific modulations influence the output---thereby validating the effectiveness of our hierarchical spatial control.
% Furthermore, Fig. \ref{fig:animal} demonstrates the model's capacity for intra-class diversity, showing how stochastic sampling with different seeds yields visually distinct entities under a broad concept.
\label{sec:more_results}

\section{Broader Societal Impacts}
\label{sec:impact}

The development of \textbf{SCDiff} offers significant potential for advancing creative industries, yet it also necessitates a careful consideration of its broader societal implications.

\paragraph{Positive Impacts.} Our framework provides an efficient and controllable tool for digital artists, industrial designers, and the advertising industry. By enabling precise geometric and structural reorganization while maintaining background consistency, SCDiff reduces the technical barrier for high-quality creative synthesis. This empowers individual creators to explore complex design concepts that were previously resource-intensive, thereby fostering democratic access to advanced generative AI tools.

\paragraph{Negative Impacts and Mitigations.} Like most generative technologies, SCDiff could potentially be misused to create deceptive visual content or misleading product designs. However, we mitigate these risks through two key strategies:

\textbf{User-in-the-loop Design:} Unlike fully autonomous generators, our method is designed as a user-guided enhancement tool. By requiring explicit spatial constraints ($R$) and semantic prompts, the responsibility of intent remains with the human operator, making it less suitable for large-scale, automated disinformation campaigns.

\textbf{Deployment Recommendations:} We strongly advocate for the integration of invisible digital watermarking (e.g., SynthID) and adherence to content provenance standards (e.g., C2PA) when deploying SCDiff in commercial pipelines. These measures ensure that generated creative content can be reliably traced back to its AI-assisted origin, preventing the masquerading of synthetic designs as authentic photographs.
Overall, we believe the benefits of localized creative control significantly outweigh the risks, provided the technology is deployed within a framework of transparency and accountability.

%%%%%%%%%%%%%%%%%%%%%%%%%%%%%%%%%%%%%%%%%%%%%%%%%%%%%%%%%%%%

% \clearpage
% \newpage
% \input{checklist.tex}
\end{document}